\documentclass{article}

\usepackage{iclr2027_conference,times}
\usepackage{amsmath,amssymb,mathtools}
\usepackage{graphicx}
\usepackage{booktabs}
\usepackage{microtype}
\usepackage{xcolor}
\usepackage{url}
\usepackage{hyperref}
\usepackage{enumitem}
\usepackage{siunitx}
\usepackage{subcaption}

\hypersetup{colorlinks=true,citecolor=blue,linkcolor=blue,urlcolor=blue}
\graphicspath{{figures/}}

\newcommand{\trace}{\textsc{TRACE}}
\newcommand{\real}{\mathrm{real}}
\newcommand{\twin}{\mathrm{twin}}
\newcommand{\diff}{\mathrm{diff}}
\newcommand{\R}{\mathcal{R}}
\newcommand{\C}{\mathcal{C}}
\newcommand{\E}{\mathcal{E}}
\newcommand{\vect}[1]{\boldsymbol{#1}}
\newcommand{\dphat}[1]{\widehat{\Delta\vect{p}}_{#1}}   
\newcommand{\B}{\mathcal{B}}
\newcommand{\agg}{\mathrm{agg}}
\newcommand{\bld}{\mathrm{bld}}
\newcommand{\fuse}{\mathrm{fuse}}
\newcommand{\probe}{\mathrm{probe}}

\iclrfinalcopy

\title{\trace: Learning to Repair Wireless Digital Twins from Radio Residuals}

\title{\trace{}: Learning to Self-Calibrate Wireless Digital Twins from ISAC Measurements}

\title{\trace{}: Learning to Self-Calibrate Wireless Digital Twins from Radio Residuals}

\author{Saad Masrur$^1$, Saeed R. Khosravirad$^2$, \.{I}smail G\"{u}ven\c{c}$^1$\\
$^1$Department of Electrical and Computer Engineering, North Carolina State University, Raleigh, NC\\
$^2$Radio Systems Research, Nokia Bell Labs, Murray Hill, NJ, USA\\
{\tt \{smasrur,iguvenc\}@ncsu.edu}
}

\begin{document}
\maketitle
\lhead{Preprint. Under review.}
\begin{abstract}
Wireless digital twins (DTs) rely on 3D environment models to predict radio propagation and support wireless-network decisions, yet these models are often initialized from imperfect 3D maps. Errors in building position, height, footprint, and orientation can therefore cause a high-fidelity propagation engine to simulate the wrong physical environment. In this paper, we study how a deployed wireless network can repair an existing DT using its own radio frequency (RF) measurements. In particular, we introduce \textbf{Twin Residual Alignment and Calibration Engine} \textbf{(\trace{})}, a physics-grounded learning-based self-calibration framework
that treats twin maintenance as residual alignment between the physical world
and the current DT. Using the same sensor geometry and radio configuration as the physical measurements, \trace{} uses ray tracing to generate corresponding RF observations from the current DT and compares them with the measured RF observations. The measured and simulated RF data are coherently backprojected onto a common metric world grid. For each building, \trace{} extracts the same local region from both backprojections, centered at the building position specified by the current DT. \trace{} then learns a multi-view corrector that combines evidence from multiple sensing nodes, accounts for their viewing geometry and the relative arrangement of buildings, and predicts a gated six-parameter geometric correction for each building. A key advantage of the \trace{} design is that it suppresses shortcuts tied to absolute layout and sensor ordering, supports variable numbers of buildings and sensing nodes, and enables iterative correction through re-rendering. Across 5,400 held-out samples from unseen simulated scenes at \SI{28}{GHz}, \trace{} reduces 3D position RMSE from \SI{2.202}{m} to \SI{0.302}{m} and yaw RMSE from $4.978^\circ$ to $0.894^\circ$. It also outperforms ViT and U-Net baselines under changes in layout, building count, sensing-node count, and SNR. On measured \SI{28}{GHz} RF data from the NIST outdoor courtyard, a model trained only on synthetic RF reduces mean planar wall-position error from \SI{1.00}{m} to \SI{7.8}{cm}, requiring neither measured-data fine-tuning nor ground-truth geometric labels for the measured data. These results show that the discrepancy between measured and twin-rendered RF observations can serve as a learning signal for repairing a wireless DT.
\end{abstract}

\section{Introduction}
\label{sec:intro}


\begin{figure}[t]
    \centering
    \includegraphics[width=0.99\textwidth]{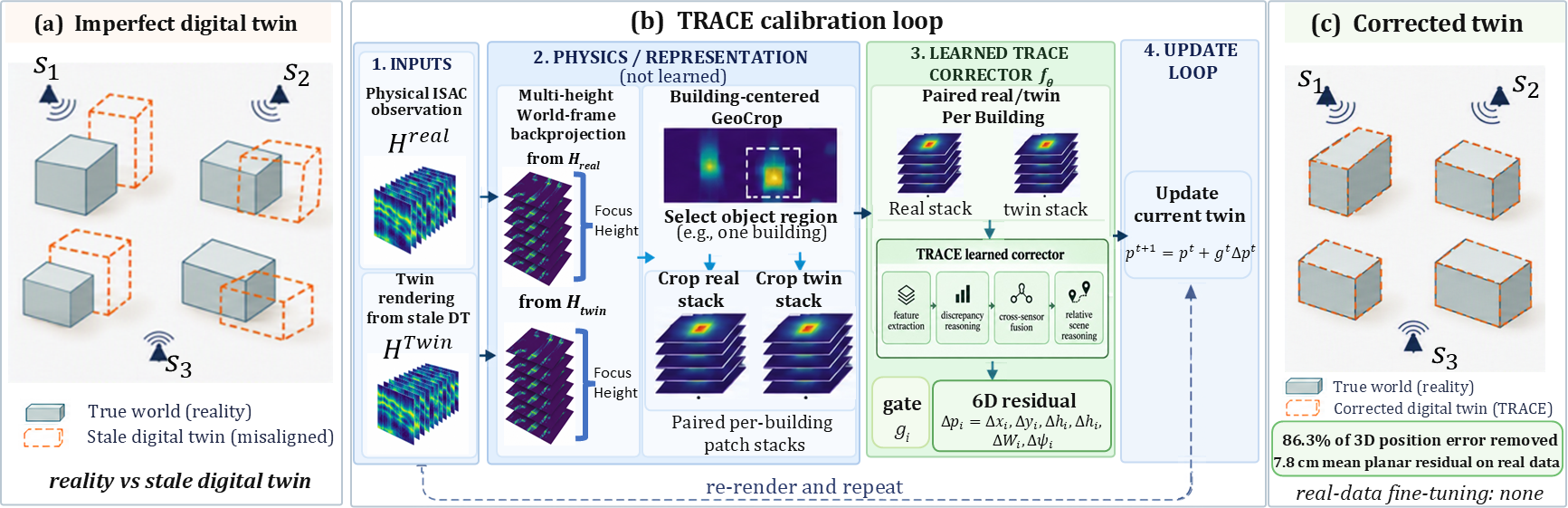}
    \caption{\textbf{\trace{} at a glance.} (a) A twin with wrong building geometry. (b) Network and twin observe the same acquisition; both are backprojected into one world frame, and GeoCrop makes each building a local alignment problem; the corrector emits a gate and 6D residual, written back and re-rendered. (c) 86.3\% of simulated 3D position error removed; \SI{7.8}{cm} on measured NIST RF.}
    \label{fig:overview}
\end{figure}


Wireless learning systems increasingly incorporate representations of the physical environment into their predictions, whether as maps, learned scene fields, or 3D point-cloud or mesh geometry~\citep{levie2021radiounet,zhao2023nerf2,zhang2026radtwin,hehn2025wigatr}. These representations provide the spatial context that determines how radio signals propagate and influence downstream predictions and decisions. Prediction accuracy therefore depends on the fidelity of the environmental representation, which is often derived from imperfect maps and may not match the physical scene.


Outdoor twins are extruded from public maps and are prone to non-negligible error. \citet{fan2014osm} report a \SI{4.13}{m} mean offset between OpenStreetMap (OSM) footprints and authoritative reference data, and of 507 million OSM buildings, only 2.9\% carry a height tag~\citep{biljecki2023osm}. Geometric errors can distort key wireless behavior, including signal strength, delay, and link reliability, more strongly than errors in material properties~\citep{tadik2025opengert}, and at that scale a corrupted twin keeps only 14\% of oracle beam-selection performance on the weakest links (Sec.~\ref{sec:results_real}). 




\textbf{A key question remains: how can a deployed wireless network repair the twin it already has?} Existing work takes two routes: training forward predictors to tolerate inaccuracies in the 3D environment model~\citep{jaensch2026noisyenv}, without updating the twin, or inferring geometric structure from RF, such as scatterers or object shape and pose~\citep{hehn2025wigatr,luo2025sgr,chen2026rfdt}, rather than correcting an existing 3D scene model. This leaves a different practical setting: an operator holds an approximate twin with known buildings and needs to know which parts are wrong. We call this \textbf{residual twin maintenance}, distinct from robustness, reconstruction, and per-instance inverse rendering. The network acquires an integrated sensing and communication (ISAC) observation~\citep{liu2022isac} while the twin renders the same acquisition using RF ray tracing, and their disagreement provides the evidence from which the geometric error is inferred (Sec.~\ref{sec:problem}).

\textbf{The central challenge is an RF representation from which geometric errors can be inferred and corrected across unseen scenes.}
Raw CSI entangles contributions from the entire scene, encouraging
layout-specific rather than transferable correction rules. \trace{} addresses this with two complementary operations. First, coherent backprojection places measured and twin-rendered RF in a common metric world frame. Second, for each building, it extracts matched local regions around its current twin location, a building-centered operation we call \textit{GeoCrop}. Together, backprojection and GeoCrop transform a scene-wide RF mismatch into a localized
measured-to-twin comparison that generalizes across unseen layouts
(Sec.~\ref{sec:results_representation}).



We make four contributions:
\begin{itemize}[leftmargin=1.15em,nosep]
    \item \textbf{Residual calibration of an existing digital twin (DT).}
    We formulate twin maintenance as selective correction of an existing building-level 3D prior, rather than reconstructing geometry from scratch or optimizing a new scene for each deployment.
    \item \textbf{A building-centered representation for transferable correction.} We combine multi-height world-frame backprojection with GeoCrop to convert global RF observations into local residual queries tied to the twin, enabling transfer across unseen layouts and varying building counts.
    \item \textbf{A learned multi-view geometric corrector.} We design a pose-aware multi-view corrector that compares measured and twin-rendered RF, fuses complementary sensing views, and uses relative building geometry to predict gated six-parameter corrections without fixed sensor or building slots.
    \item \textbf{Evaluation across distribution shifts and measured RF.}
    We test unseen layouts, building counts, geometric-error ranges, sensing-node subsets, and SNR levels, together with falsification and iterative correction. We further demonstrate transfer to measured \SI{28}{GHz} NIST RF data without measured-data fine-tuning and evaluate the corrected twin on downstream beam selection.
\end{itemize}

\section{Related Work}
\label{sec:related}

\textbf{Geometry is the conditioning state.}
RadioUNet predicts urban path loss from building maps~\citep{levie2021radiounet}, and later radio-environment-map methods retain this geometry-conditioned formulation~\citep{zadeh2026rem,jaensch2026noisyenv}. DeepMIMO likewise derives channels from a site-specific 3D environment~\citep{alkhateeb2019deepmimo}, while NeRF$^2$ and SpecNeRF encode scene propagation implicitly in learned fields~\citep{zhao2023nerf2,basu2024specnerf}. More recent models expose geometry explicitly through learned electromagnetic properties, ray--surface interactions, point clouds, or equivariant mesh representations~\citep{jiang2025lwdt,orekondy2023winert,zhang2026radtwin,hehn2025wigatr}. Across these approaches, the environmental state is assumed to be sufficiently accurate; \trace{} instead asks how to correct it when the existing twin is wrong.
\paragraph{Geometry recovery and residual twin correction.} RF measurements can be used to infer scene structure: NeRF-style methods and Wi-GATr recover geometry conditioned on wireless observations, CSI methods reconstruct scatterers, and RFDT optimizes object geometry against measured RF through differentiable rendering~\citep{zhao2023nerf2,hehn2025wigatr,luo2025sgr,bazzi2025isacimaging,chen2026rfdt}. Sionna RT gradients ignore visibility changes~\citep{hoydis2023sionna}, limiting gradient-based correction of buildings. These are closely related inverse problems, but \trace{} starts from a different prior: an approximate building-level twin with known correspondences. Its own RF rendering serves as the hypothesis to correct, and the model predicts only the residual update to that existing geometry rather than recovering it from scratch. TRACE learns a reusable residual corrector, rather than solving a new optimization problem for each deployment. Related calibration work on phase and material parameters is complementary~\citep{ruah2023phasecal,kang2026vlm}.


\paragraph{Why geometry correction from RF is difficult.} The RF signature of a geometric error depends strongly on sensing viewpoint. For a building near the boresight of a sensing node, lateral displacement changes the propagation distance weakly and is therefore difficult to infer from a single view, whereas radial displacement appears more directly in delay. 
Transverse motion is expressed more strongly through phase and angular structure, motivating phase-preserving processing and fusion across separated sensing nodes with complementary viewpoints. 
The residual is subtle: even with building displacements up to \SI{4}{m}, the physical and stale-twin channels remain 96.5\% coherent, and the largest elementwise change is only 0.46\% of the channel peak.  Thus, the corrector must recover a small, viewpoint-dependent geometric mismatch rather than distinguish an entirely different scene.

\section{Residual Twin Maintenance}
\label{sec:problem}

Let the current twin contain $N$ rectangular 3D buildings. Building $i$ is represented by
$\vect{p}_i^{(t)}=[x_i,y_i,h_i,\ell_i,w_i,\psi_i]$, where $(x_i,y_i)$ denote the center of the building, $h_i$ its height, $(\ell_i,w_i)$ its footprint dimensions, and $\psi_i$ its yaw. The complete twin state is $\vect{P}^{(t)}=\{\vect{p}_i^{(t)}\}_{i=1}^{N}$, where $t$ indexes the current correction iteration. We represent the physical scene in the same form as $\vect{P}^{\star}=\{\vect{p}_i^{\star}\}_{i=1}^{N}$. Rather than reconstructing $\vect{P}^{\star}$ from scratch, we estimate for each building the correction $\Delta\vect{p}_i^{(t)}=\vect{p}_i^{\star}-\vect{p}_i^{(t)}$ needed to align the current twin with the physical scene.


Let \(S\) denote the number of sensing nodes. In the ISAC setting, these are wireless base stations that provide communication while also sensing the environment~\citep{liu2022isac}. For each node \(s\in\{1,\ldots,S\}\), let \(\vect{\xi}_s\) denote its known sensing configuration, including pose, waveform, bandwidth, and array geometry. The node measures complex CSI \(\vect{H}^{\real}_{s}\), while the current twin simulates the corresponding CSI under the same configuration using an RF ray tracer such as Sionna RT~\citep{hoydis2023sionna},
\(\vect{H}^{\twin}_{s}=\R(\vect{P}^{(t)};\vect{\xi}_s).\)
Using the same \(\vect{\xi}_s\) makes the two RF responses directly comparable, so their mismatch provides the signal for geometric correction. Receiver noise is included in the physical branch; the CSI representation and noise treatment are detailed in App.~\ref{app:formulation}.

We coherently backproject $\vect{H}^{\real}_{s}$ and $\vect{H}^{\twin}_{s}$ onto a shared world grid at $K$ focus heights, yielding stacks of world-frame RF maps $\vect{I}^{\real}$ and $\vect{I}^{\twin}$ (Sec.~\ref{sec:frontend}).This places RF evidence from all sensing nodes in the same physical coordinate system, so a building appears at the same world location regardless of which node observes it (App.~\ref{app:formulation}). From $(\vect{I}^{\real},\vect{I}^{\twin},\vect{P}^{(t)})$, the learned
corrector predicts a gate $g_i\in[0,1]$ and raw residual $\vect{r}_i$.
The applied correction is
$\widehat{\Delta\vect{p}}_i=g_i\vect{r}_i$, giving
\begin{equation}
    \vect{p}_i^{(t+1)}
=
\vect{p}_i^{(t)}+\widehat{\Delta\vect{p}}_i,
    \qquad
    \vect{P}^{(t+1)}
    =
    \vect{P}^{(t)}
    +
    f_{\theta}\!\left(
    \vect{I}^{\real},
    \vect{I}^{\twin},
    \vect{P}^{(t)}
    \right).
    \label{eq:loop}
\end{equation}

The corrected twin is then re-rendered, allowing the same correction rule to be applied iteratively. Because each prediction becomes the next input, avoiding scene- or sensor-specific shortcuts is important for stable closed-loop correction (Sec.~\ref{sec:results_invariance}).



\begin{figure}[t]
    \vspace{-1mm}
    \centering
    \includegraphics[width=0.99\textwidth]{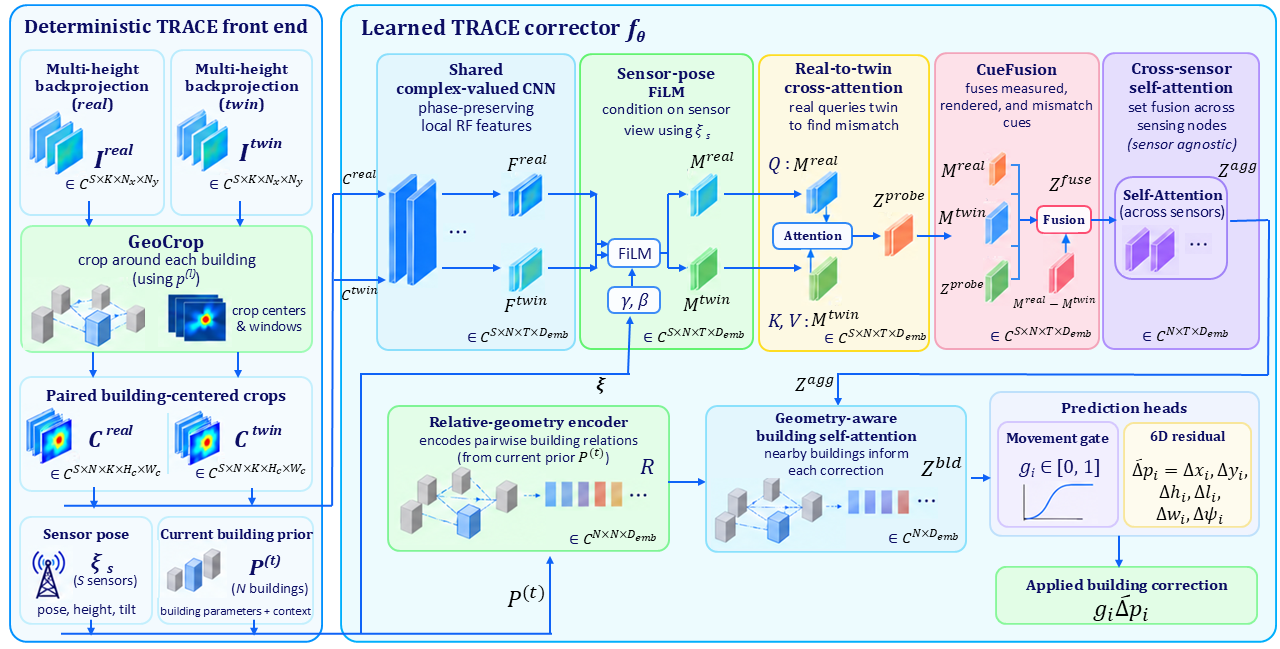}
    \caption{\textbf{\trace{} architecture: from RF alignment to twin correction.}
    The front end aligns measured and twin-rendered RF in a common world frame and extracts paired GeoCrops. The corrector compares the two RF branches, fuses sensing-node and neighboring-building context, and predicts a gated six-parameter geometric correction.}
    \label{fig:corrector}
\end{figure}

\section{\trace}
\label{sec:method}



Fig.~\ref{fig:corrector} summarizes \trace{}, which consists of a deterministic RF representation stage followed by a learned corrector $f_\theta$. The deterministic stage aligns measured and twin-rendered RF in a common world frame and forms paired GeoCrops per building, which the corrector maps to gated six-parameter geometric updates. Only $f_\theta$ is learned; ray tracing, backprojection, and GeoCrop are deterministic.

\subsection{Deterministic front end}
\label{sec:frontend}


\paragraph{Coherent world-frame backprojection registers the RF observations.}
Raw CSI mixes contributions from the entire scene in one antenna--frequency tensor, so the RF response of an individual building is not spatially isolated. Range--angle maps provide spatial separation, but they are expressed in each sensing node's local frame; the same building therefore appears at different ranges and angles across nodes. To place all observations in a common physical coordinate system, \trace{} uses coherent backprojection. Backprojection is a spatial focusing operation: for each candidate world point $\vect{q}=(x,y,z)$, it compensates each complex RF measurement for the propagation delay expected from that point and then sums the compensated measurements. If $\vect{q}$ corresponds to a true reflecting location, the phases align and add coherently; otherwise they tend to cancel. Formally,
\begin{equation}
 I_s(\vect{q})
 =
 \sum_{a,k}
 w_{a,k}(\vect{q})\,H_{s,a,k}\,
 e^{j2\pi f_k\tau_{s,a}(\vect{q})},
 \label{eq:bp}
\end{equation}
where $a$ indexes a transmit--receive antenna pair, $k$ the subcarrier, $H_{s,a,k}$ the corresponding complex CSI coefficient, $\tau_{s,a}(\vect{q})$ the expected propagation delay, and $w_{a,k}(\vect{q})$ the backprojection weight. Applying the same operation to the measured and twin-simulated CSI places both branches and all sensing nodes on a common metric world grid. Repeating it at $K$ focus heights preserves height-dependent structure and complex phase information. Further details are in Apps.~\ref{app:formulation} and~\ref{app:complex}.

\paragraph{GeoCrop converts global RF evidence into local residual alignment.} The full-scene backprojected RF maps expose absolute location, letting the model learn layout-specific patterns. For each building $i$ and sensing node $s$, GeoCrop uses the building position $(x_i^{(t)},y_i^{(t)})$ from the \emph{current twin} as the center and crops a fixed $H_p\times W_p$ window around it from the backprojection. The same window is cropped from measured and twin-simulated backprojections at all $K$ heights; we never use the unknown true position. If the current twin geometry is correct, the two local RF structures align. If the building is displaced, resized, or rotated, the resulting mismatch appears directly within the same local frame. The corrector thus learns building updates from local RF mismatch, not the full scene. App.~\ref{app:scenebp} and Fig.~\ref{fig:scene_bp} illustrate world-frame backprojection and GeoCrop across sensing nodes.

\subsection{The learned corrector: four design decisions}
\label{sec:corrector}

The learned corrector follows four design steps: preserving viewpoint-dependent RF cues, comparing measurements with the current twin hypothesis, combining complementary sensing views, and introducing scene context without exposing absolute building location; see details in App.~\ref{app:arch}.



\paragraph{(i) Shared complex encoding with sensor-pose FiLM.}
For each building and sensing node, the corrector uses three aligned GeoCrops: measured, rendered, and a coherent residual formed from their complex world-frame difference before cropping (App.~\ref{app:paired_inputs}). A shared complex-valued CNN (CV-CNN) converts each crop into tokens while preserving amplitude and phase (App.~\ref{app:complex_encoder}). Weight sharing lets the same encoder handle varying numbers of buildings and sensing nodes. Because geometric errors produce viewpoint-dependent RF signatures, sensor-pose feature-wise linear modulation (FiLM) conditions the tokens on the node's position and orientation (App.~\ref{app:film}).


\paragraph{(ii) Real-to-twin cross-attention and CueFusion.}

This step compares, for each building and sensing node, the measured RF with the RF that the \emph{current twin} predicts. Measured tokens therefore query the corresponding rendered tokens, directly comparing the physical observation with the current twin hypothesis before information is mixed across nodes or buildings (App.~\ref{app:real_twin_attn}). CueFusion then combines this cross-attended representation with the measured and rendered features, their feature-space difference, and the encoded coherent RF difference (App.~\ref{app:cuefusion}).



\paragraph{(iii) Permutation-invariant cross-sensor fusion.}
Different sensing nodes provide complementary views of the same building; one may clearly observe a structure while another is weak or occluded. \trace{} therefore attends across the discrepancy tokens from all nodes and pools them into a single building representation (App.~\ref{app:sensor_fusion}). The fusion treats sensing nodes as an unordered set, allowing the corrector to operate with different node subsets and orderings (App.~\ref{app:ablation_permutation}).

\paragraph{(iv) Geometry-aware building attention.}

After cross-sensor fusion, each building is represented by one RF-discrepancy token. However, buildings are not independent: their relative arrangement provides geometric context for interpreting whether and how a particular building should be corrected. \trace{} therefore augments each token with the building's current shape $(h,\ell,w,\psi)$ from the twin (App.~\ref{app:shape_prior}) and applies geometry-aware self-attention across buildings. The attention between buildings $i$ and $j$ is conditioned on their relative geometry, including relative position, height, dimensions, distance, and orientation (App.~\ref{app:building_attention}). Absolute $(x_i,y_i)$ coordinates are excluded, so the interaction depends on relative building arrangement rather than global map location.

The resulting token for each building is passed to two heads (App.~\ref{app:prediction_heads}): a gate $g_i$ determines whether the building should be updated, and a regression head predicts its six-parameter residual $(\Delta x,\Delta y,\Delta h,\Delta\ell,\Delta w,\Delta\psi)$. The gate suppresses unnecessary corrections when the measured RF is already consistent with the current twin.


\subsection{Gated residual head and training}
\label{sec:gate}

For training, the corrupted twin and true geometry are both known, providing the target residual $\Delta\vect{p}^{\star}$. The predicted correction is $\widehat{\Delta\vect{p}}_i=g_i\,\vect{r}_i$, with gate $g_i$ and raw residual $\vect{r}_i$. The MSE supervises the residual correction, while the binary cross-entropy (BCE) supervises the update gate. We optimize:
\begin{equation}
\mathcal{L}=\frac{1}{6N}\sum_{i=1}^{N}\left\|\widehat{\Delta\vect{p}}_i-\Delta\vect{p}_i^{\star}\right\|_2^2-\frac{\lambda_{\rm gate}}{N}\sum_{i=1}^{N}\left[g_i^{\star}\log g_i+(1-g_i^{\star})\log(1-g_i)\right]
\label{eq:loss}
\end{equation}


\section{Experimental Setup}
\label{sec:experiments}

\paragraph{Simulation benchmark.}
We construct a physically rendered \SI{28}{GHz} benchmark in Sionna RT with three separated rooftop monostatic sensing nodes \(S=3\), $8\times8$ arrays, and \SI{122.88}{MHz} bandwidth. Dataset generation separates \emph{physical-scene variation} from \emph{twin error}. First, we generate diverse physical scenes as $\vect{p}_i^\star=\vect{p}_i^{\rm nom}+\vect{\eta}_i$ by independently varying each building's position, height, footprint dimensions, and yaw. We then create a corrupted twin as $\vect{p}_i^{(0)}=\vect{p}_i^\star+\vect{\varepsilon}_i$ for a subset of buildings, with position errors up to \SI{4}{m} together with errors in height, footprint, and orientation, while leaving the remaining buildings unchanged. Thus, $\vect{\eta}_i$ controls scene diversity, whereas $\vect{\varepsilon}_i$ defines the geometric error to be corrected. We ray-trace the physical scene and the corrupted twin under the same sensing configuration, with receiver noise added only to the physical branch (App.~\ref{app:data}). Evaluation uses 5,400 scene-disjoint samples and tests generalization without retraining across building counts (4, 8, and 10 versus 6 in training), SNR from \SIrange{-5}{20}{dB}, unseen layouts, geometric errors outside the training corruption range, and changed sensing-node poses. \textbf{Out-of-distribution (OOD)-Layout} places $\vect{\eta}_i$ outside the training location support, while \textbf{OOD-Error} increases $\vect{\varepsilon}_i$ beyond the training corruption range.

\paragraph{Measured benchmark and sim-to-real transfer.} 
We next evaluate \trace{} using measured \SI{28}{GHz} RF data from the NIST outdoor courtyard, obtained from the NextG Channel Model Alliance data repository~\citep{nist_nextg_repository}. This tests whether a corrector trained entirely on synthetic twin data can transfer directly to real measured RF without fine-tuning. The setup uses one fixed transmitter and a mobile receiver, and each TRACE estimate fuses RF measurements from $M=8$ receiver positions in Area~1, as illustrated in Fig.~\ref{fig:nist_measured_data} and detailed in App.~\ref{app:nist}. From the LiDAR-derived reference geometry, we construct a Sionna DT of the courtyard. \trace{} is trained entirely on synthetic RF pairs generated from the courtyard
twin and its randomized corruptions; no measured RF data is used for training. At test time, the physical branch is replaced by measured NIST RF, while the corresponding corrupted twin is rendered in Sionna under the same acquisition configuration. The frozen \trace{} model then uses this measured-to-rendered RF discrepancy to predict the planar wall corrections. LiDAR is used only as the geometric reference for constructing the twin and evaluating the recovered wall positions.


\paragraph{Baselines, metrics, and residual scope.}
We use the corrupted twin itself as the no-correction reference and compare \trace{} against full-scene ViT~\citep{dosovitskiy2021image} and U-Net ~\citep{ronneberger2015unet} baselines trained under the same splits, noise model, targets, loss, and schedule (App.~\ref{app:baselines}). We report per-parameter RMSE and the percentage of initial error removed.


\section{Results}
\label{sec:results}

\subsection{The RF Representation Design and Correction Performance}
\label{sec:results_representation}

\begin{table}[t]
\begin{minipage}[t]{0.505\textwidth}
\centering
\caption{\textbf{Effect of RF representation on layout generalization.}
All rows use the same \trace{} model; only the RF input representation changes. Error removed is relative to the corrupted twin before correction (\SI{1.878}{m} 2D RMSE); ``+ GeoCrop'' is the full \trace{} representation.}
\label{tab:repr}
\footnotesize
\setlength{\tabcolsep}{2.5pt}
\begin{tabular}{lcc|cc}
\toprule
& \multicolumn{2}{c|}{\textbf{Standard}} & \multicolumn{2}{c}{\textbf{OOD-Layout}} \\
\textbf{Input} & RMSE & Rem. & RMSE & Rem. \\
\midrule
Raw CSI          & 0.255 & 86.4\% & 2.111 & $-12.4$\% \\
World-frame BP   & \textbf{0.186} & \textbf{90.1\%} & 0.824 & 56.1\% \\
\ + GeoCrop      & 0.205 & 89.1\% & \textbf{0.213} & \textbf{88.7\%} \\
\bottomrule
\end{tabular}
\end{minipage}\hfill
\begin{minipage}[t]{0.475\textwidth}
\centering
\caption{\textbf{Full six-parameter geometric correction.}
RMSE on scene-disjoint test layouts for the corrupted twin, full-scene ViT and
U-Net baselines, and \trace{}; lower is better.}
\label{tab:main}
\footnotesize
\setlength{\tabcolsep}{2.5pt}
\begin{tabular}{lccccc}
\toprule
\textbf{Method} & \textbf{2D} & \textbf{H} & \textbf{L} & \textbf{W} & \textbf{Yaw} \\
& [m] & [m] & [m] & [m] & [$^\circ$] \\
\midrule
Corrupted twin & 1.878 & 1.151 & 0.871 & 0.868 & 4.978 \\
ViT            & 1.483 & 0.651 & 0.727 & 0.727 & 5.050 \\
U-Net          & 0.553 & 0.640 & 0.659 & 0.663 & 3.702 \\
\textbf{\trace} & \textbf{0.205} & \textbf{0.221} & \textbf{0.273} & \textbf{0.260} & \textbf{0.894} \\
\bottomrule
\end{tabular}
\end{minipage}
\end{table}

\textbf{Rich RF information alone does not guarantee transfer.}
Table~\ref{tab:repr} compares raw CSI, world-frame BP, and BP+GeoCrop while keeping the correction task fixed. Raw CSI contains rich geometric information, but paths from all buildings are entangled in the same antenna--frequency tensor with no explicit spatial correspondence. This allows the model to exploit and effectively memorize layout-specific RF patterns rather than learn a transferable relation between RF mismatch and geometric error. It therefore performs well on standard scenes, removing 86.4\% of the 2D position error, but fails when building layouts move outside the training distribution: OOD-Layout RMSE rises to \SI{2.111}{m}, 12.4\% worse than applying no correction. The information is present, but its raw representation encourages the wrong shortcut.

World-frame BP provides the first step by registering measured and rendered RF in a common metric coordinate system, reducing OOD-Layout RMSE to \SI{0.824}{m}. However, the complete BP image still exposes absolute building locations and global scene structure, leaving the model vulnerable to layout-specific correlations. GeoCrop removes this shortcut by extracting matched building-centered crops from the measured and rendered RF maps around each building's current twin location. As a result, performance remains nearly unchanged from standard to OOD-Layout: \SI{0.205}{m} versus \SI{0.213}{m}, with 89.1\% and 88.7\% of the initial error removed. Together, BP provides physical registration, and GeoCrop makes that registered residual transferable across layouts.

\textbf{\trace{} converts local RF evidence into accurate full-geometry correction.}
We next compare the complete \trace{} system with full-scene ViT and U-Net baselines on the standard scene-disjoint test set (Table~\ref{tab:main}). Both baselines process the entire world-frame BP map and compress it into a global representation that is reused for every building. They must therefore infer which spatial RF discrepancy corresponds to each building. In contrast, \trace{} uses GeoCrop to extract a building-specific BP region, then the learned corrector compares the two branches, fuses complementary sensing views, and uses relative-geometry attention to incorporate neighboring-building context before predicting the residual. \trace{} achieves the lowest error for every geometric parameter, reducing 3D position RMSE from \SI{2.202}{m} to \SI{0.302}{m} (86.3\% removed) and yaw from $4.978^\circ$ to $0.894^\circ$. Relative to the stronger U-Net, it further reduces residual RMSE by 62.9\% in position, 65.5\% in height, and 75.9\% in yaw. App.~\ref{app:ablation} isolates the contribution of the learned corrector.




\textbf{Comparison with classical registration.}
 We also test whether the planar correction (i.e., 2D) can be recovered directly from paired GeoCrops using training-free phase correlation or normalized cross-correlation (NCC). Phase correlation changes 2D RMSE from \SI{1.878}{m} to \SI{1.925}{m} ($-2.5\%$), while NCC reaches \SI{1.802}{m} ($4.0\%$), compared with \SI{0.206}{m} ($89.0\%$) for \trace{}. Thus, the building-centered RF discrepancy is not a simple rigid translation; \trace{} instead learns its mapping to the geometric correction: full details are in App.~\ref{app:classical_registration}.

\subsection{The corrector uses the intended physical evidence}
\label{sec:results_falsification}

\begin{table}[t]
\centering
\caption{\textbf{Falsifying the mechanism on the frozen model.} 3D position RMSE when RF branches are blinded (set to zero before GeoCrop) or replaced by those of a different scene, with the twin prior $\vect{P}^{(t)}$ unchanged. The uncorrected twin is \SI{2.202}{m}; $^{\dagger}$worse than no correction.}
\label{tab:falsify}
\footnotesize
\setlength{\tabcolsep}{4pt}
\renewcommand{\arraystretch}{1.15}
\begin{tabular}{l|c|ccc|cc}
\toprule
& & \multicolumn{3}{c|}{\textbf{Input blinding}} & \multicolumn{2}{c}{\textbf{Scene shuffling}} \\
& \begin{tabular}[b]{@{}c@{}}\textbf{\trace{}}\\ $(\vect{I}^{\real},\vect{I}^{\twin})$\end{tabular}
& $\vect{I}^{\real}$ \textbf{only} & $\vect{I}^{\twin}$ \textbf{only} & \textbf{Both blinded}
& \textbf{Shuffled} $\vect{I}^{\twin}$ & \textbf{Shuffled} $\vect{I}^{\real}$ \\
\midrule
3D RMSE [m] & \textbf{0.302} & 0.835 & 2.630$^{\dagger}$ & 2.460$^{\dagger}$ & 0.889 & 3.723$^{\dagger}$ \\
\bottomrule
\end{tabular}
\end{table}

\textbf{Testing what drives the twin correction.}
High accuracy alone cannot rule out that \trace{} memorizes the training corruption statistics rather than inferring each correction from RF evidence. We therefore freeze the model and selectively blind its RF inputs, with the results shown in Table~\ref{tab:falsify}. With both branches, 3D position RMSE is \SI{0.302}{m}. With only $\vect{I}^{\real}$, it rises to \SI{0.835}{m}: the measurement still reveals the physical structure, but without $\vect{I}^{\twin}$ the model loses the twin reference, showing how the current hypothesis differs from it. With only $\vect{I}^{\twin}$, RMSE reaches \SI{2.63}{m}; the model sees what the corrupted twin looks like but has no physical evidence revealing how it differs from reality. When both RF branches are blinded, the model still receives the corrupted geometry prior $\vect{P}^{(t)}$, yet RMSE remains \SI{2.46}{m}, worse than the \SI{2.202}{m} no-correction reference. Together, these results show that the measurement reveals the mismatch, and the current twin provides the reference to correct it.



\textbf{The two RF branches must also correspond to the same scene.}
For each sample, the target consists of a physical scene and its corrupted twin. We keep the twin state $\vect{P}^{(t)}$ fixed and replace one RF branch with the corresponding RF from a different scene, with the result reported in Table~\ref{tab:falsify}. When only $\vect{I}^{\twin}$ is shuffled, the correct $\vect{I}^{\real}$ still describes the physical scene being repaired, so partial correction remains, and RMSE rises to \SI{0.889}{m}. When $\vect{I}^{\real}$ is shuffled instead, the model receives physical evidence from the wrong scene, and RMSE rises to \SI{3.723}{m}, worse than no correction. Thus, \trace{} requires the measured RF and twin-rendered RF to describe the same environment.



\subsection{\trace{} Generalization Across Scene Complexity and Sensing Conditions } \label{sec:results_invariance}

\begin{figure}[t]
\vspace{-2mm}
    \centering
    \includegraphics[width=0.97\textwidth]{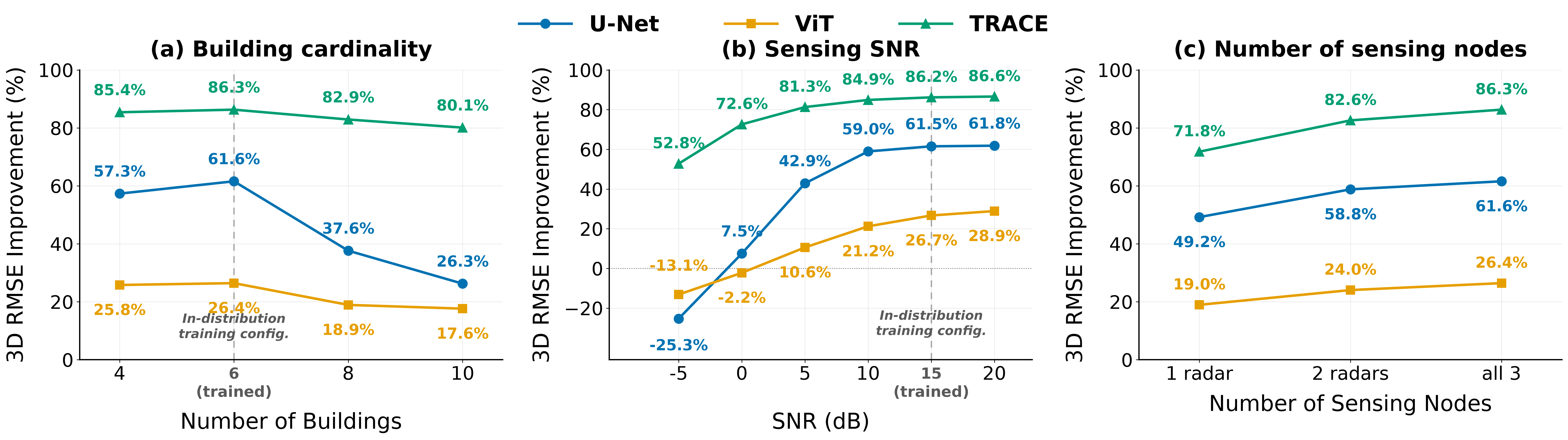}
    \caption{\textbf{Three shifts, one trained model, no retraining.} 3D RMSE improvement (\% of initial error removed) as (a) building count, (b) SNR, and (c) active sensing nodes move away from training.}
    \label{fig:invariance}
\end{figure}

\textbf{\trace{} remains stable as the number of buildings changes.}
Fig.~\ref{fig:invariance}a shows performance when models trained on six-building scenes are evaluated without retraining on four, eight, and ten buildings. \trace{} removes 85.4\%, 86.3\%, 82.9\%, and 80.1\% of the initial error, while both full-scene
baselines degrade substantially. The difference follows from how the models represent the scene. For the full-scene baselines, changing the number of buildings changes the global RF pattern from which all building corrections must be disentangled. In \trace{}, every building is still presented as the same paired GeoCrop and processed by the same encoder; additional buildings simply add building tokens, which the \trace{} architecture handles as a variable-size set. The small drop at ten buildings is consistent with increased occlusion and multipath.

\textbf{\trace{} remains robust under severe sensing noise.}
Fig.~\ref{fig:invariance}b shows performance when models trained at \SI{15}{dB} are evaluated without retraining across $-\SI{5}{dB}$ to \SI{20}{dB}. \trace{} removes 86.2\% of the initial error at \SI{15}{dB}, 72.6\% at \SI{0}{dB}, and 52.8\% at $-\SI{5}{dB}$, whereas the U-Net and ViT fall to $-25.3\%$ and $-13.1\%$, worsening the corrupted twin. Three aspects of \trace{} explain robustness. First, GeoCrop restricts the input to the region around each building, avoiding noise-only areas in the full-scene representation. Second, because noise affects the measured branch, the rendered branch remains a clean reference; third, the per-building gate can suppress unreliable updates.

\textbf{\trace{} effectively combines information across sensing nodes.}
Figure~\ref{fig:invariance}c shows performance with different numbers of active sensing nodes. All models are trained with random sensor dropout, so the reduced node sets remain in-distribution. With only one node, \trace{} removes 71.8\% of the initial error, already exceeding the U-Net with all three nodes (61.6\%). From one to three nodes, \trace{} reduces the remaining error from 28.2\% to 13.7\%, compared with roughly one quarter for the U-Net and less than one tenth for the ViT. Separated nodes provide complementary viewing angles, which help resolve geometric errors that are ambiguous from a single direction. \trace{} preserves these views separately and fuses them per building, allowing the contribution of each node to depend on the RF evidence and viewing geometry.


\textbf{\trace{} corrects errors beyond the training range through iterative re-rendering.}
When every corrupted building lies outside the training error range, one pass removes 79.2\% of the initial 3D position error, versus 54.6\% for the U-Net. Re-rendering the corrected twin and applying the same frozen corrector raises this to 93.1\%. Full details are provided in App.~\ref{app:closedloop}.

\subsection{\trace{} Evaluation on Real \SI{28}{GHz} Measurements and Downstream Tasks}
\label{sec:results_real}

\begin{figure}[t]
\vspace{-1mm}
    \centering
    \includegraphics[width=0.85\textwidth]{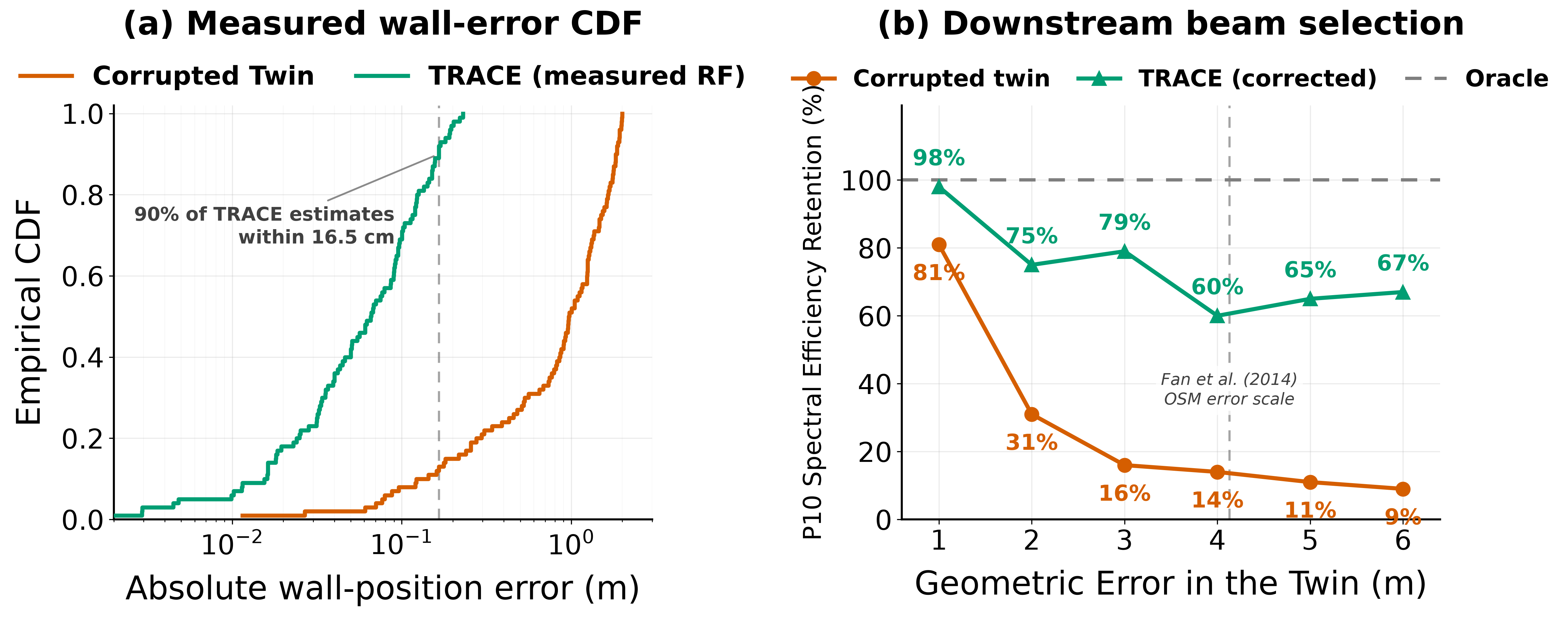}
    \caption{\textbf{Measured-RF validation and downstream impact.}
(a) Wall-position error using measured \SI{28}{GHz} NIST RF.
(b) P10 spectral-efficiency retention of beam selection using corrupted and \trace{}-corrected twins; the dashed line marks the \SI{4.13}{m} mean OSM footprint offset of \citet{fan2014osm}.}
    \label{fig:real}
\end{figure}

\textbf{\trace{} transfers to real measurements without measured-data supervision.}
Following the synthetic-only training protocol described in
Sec.~\ref{sec:experiments}, we evaluate the frozen corrector directly on the measured \SI{28}{GHz} NIST data. At test time, the physical branch is formed from the
measured response, while the twin branch is obtained by ray tracing the corrupted courtyard twin (App.~\ref{app:nist}).  Fig.~\ref{fig:real}a shows that, over 100 wall-correction samples, \trace{} reduces the mean planar error from \SI{1.00}{m} to \SI{7.8}{cm}, a 92.2\% reduction, with a \SI{6.6}{cm} median and \SI{16.5}{cm} 90th percentile. Moreover, 69\% of the estimates are within
\SI{10}{cm} and 84\% within \SI{15}{cm} of the LiDAR reference. This result demonstrates direct sim-to-real transfer: a model trained entirely in simulation performs accurate correction on real measurements without any real-data training or fine-tuning. Equally important, it also shows that \trace{} can be applied to real measurements without collecting ground-truth geometric labels for those measurements, avoiding a difficult and costly real-world labeling process. Robustness to the number of measured receiver positions $M$ is evaluated in App.~\ref{app:nist} (Fig.~\ref{fig:nist_views}).

\textbf{Correcting the twin recovers downstream beam-selection performance.}
We next ask whether improving the twin geometry also improves the wireless decision it supports. For each geometry, we select the beam that maximizes the spectral efficiency predicted by its corresponding channel. An inaccurate twin can therefore produce a different, and potentially suboptimal, beam choice, while a corrected twin should move the decision toward that obtained with the true geometry. Fig.~\ref{fig:real}b evaluates the beams selected from the corrupted and \trace{}-corrected twins on the ground-truth channel, using the beam selected from the ground-truth geometry as the oracle. We report the 10th-percentile (P10) spectral-efficiency retention relative to this oracle. As geometric error increases from \SI{1}{m} to \SI{6}{m}, retention using the corrupted twin falls from 81\% to 9\%, whereas the \trace{}-corrected twin retains 60--67\% at \SIrange{4}{6}{m}. At the \SI{4}{m} mean OSM footprint offset reported by \citet{fan2014osm}, correction raises retention from 14\% to 61\%. Thus, repairing the geometry recovers a substantial fraction of the beam-selection performance lost because of an inaccurate twin.

\section{Discussion and Limitations}
\label{sec:limitations}
\trace{} assumes that the buildings to be corrected already exist in the twin, so it cannot add a missing building or remove one that no longer exists. Our experiments also use rectangular buildings; other static elements such as bridges and vegetation, as well as dynamic objects such as vehicles and pedestrians, are left for future work.   






\section{Conclusion}
\label{sec:conclusion}


\trace{} formulates maintenance of an imperfect wireless DT as residual geometric correction between physical RF observations and the response simulated by the current twin. World-frame backprojection and GeoCrop turn the global RF mismatch into building-centered comparisons, while the learned multi-view corrector predicts gated geometric updates that generalize across layouts, building counts, sensing conditions, and error ranges. On unseen simulated scenes, \trace{} reduces 3D position RMSE from \SI{2.202}{m} to \SI{0.302}{m}, removing 86.3\% of the initial error. On measured \SI{28}{GHz} NIST RF, a model trained entirely on synthetic data reduces mean planar wall-position error from \SI{1.00}{m} to \SI{7.8}{cm} without measured-RF fine-tuning, while the corrected geometry also improves downstream beam selection. Together, these results show that a wireless network's own RF observations can provide the evidence needed to update an imperfect DT.



\subsubsection*{AI Use Statement}
Generative AI tools were used to assist with manuscript structuring,
text editing, and presentation of experimental results. They were not used to generate experimental data, numerical results, or citations. All AI-assisted content was independently verified by the authors against primary sources, experiment logs, and code. The authors take responsibility for the final content of the paper.

\subsubsection*{Reproducibility Statement}
Appendix~\ref{app:implementation} gives the simulator configuration, corruption distributions, sensing geometry, front-end parameters, architecture, hyperparameters, and evaluation protocol; Appendix~\ref{app:nist} documents the measured benchmark. 


\bibliographystyle{iclr2027_conference}
\bibliography{TRACE_ICLR2027_references_updated}

\appendix
%
%
\newpage

\section*{Overview of the Appendix}

The appendix follows the structure of the main text. Appendices~\ref{app:formulation}--\ref{app:arch} complete the method: App.~\ref{app:formulation} formalizes residual twin maintenance and the operators of the deterministic front end, App.~\ref{app:scenebp} illustrates what the corrector receives for a representative scene, and App.~\ref{app:arch} gives layer-level definitions of the learned corrector.

Appendices~\ref{app:complex}--\ref{app:nist} report the evidence behind the main claims, in the order of Sec.~\ref{sec:results_representation}--\ref{sec:results_real}. App.~\ref{app:complex} and App.~\ref{app:height_ablation} show why the representation retains complex phase and multiple focus heights; App.~\ref{app:classical_registration} shows that classical registration on the same GeoCrops does not recover the correction, so a learned corrector is needed; App.~\ref{app:ablation} isolates the contribution of each corrector component, compares it with a U-Net corrector on the same GeoCrops, and tests unseen sensing poses;   App.~\ref{app:falsification} extends the falsification tests to all geometric parameters; App.~\ref{app:closedloop} details the out-of-distribution evaluation and closed-loop correction; and App.~\ref{app:nist} documents the measured NIST benchmark. App.~\ref{app:implementation} lists the implementation, training, and baseline configurations used for all reported numbers.

\section{Problem Formulation and Notation}
\label{app:formulation}

Section~\ref{sec:problem} defines residual twin maintenance, the matched acquisition, and the gated update. This appendix supplies the objects that were stated there only in words: the tensor shapes of the two CSI branches, the backprojection and GeoCrop operators, the difference between the synthetic and measured branches, and the exact quantity the corrector is supervised on. Table~\ref{tab:notation} collects the symbols used throughout the paper.

\begin{table}[htbp]
\caption{Symbols used in the main text and appendix.}
\label{tab:notation}
\centering
\small
\setlength{\tabcolsep}{5pt}
\begin{tabular}{ll}
\toprule
Symbol & Meaning \\
\midrule
$\vect{p}_i^{(t)}=[x_i,y_i,h_i,\ell_i,w_i,\psi_i]$
  & state of building $i$ in the current twin \\
$\vect{P}^{(t)}=\{\vect{p}_i^{(t)}\}_{i=1}^{N}$
  & current twin state, $N$ buildings \\
$\vect{P}^{\star}$
  & physical scene, expressed in the same parameterization \\
$\Delta\vect{p}_i^{(t)}=\vect{p}_i^{\star}-\vect{p}_i^{(t)}$
  & target residual for building $i$ \\
$\vect{\xi}_s$
  & known acquisition configuration of sensing node $s$ \\
$\vect{H}^{\real}_{s},\ \vect{H}^{\twin}_{s}$
  & measured and twin-rendered CSI at node $s$ \\
$\R(\cdot;\vect{\xi}_s)$
  & RF ray tracer used to render the twin \\
$\B_z$
  & coherent backprojection onto the world grid at height $z$ \\
$\vect{I}^{\real}_{s},\ \vect{I}^{\twin}_{s},\ \vect{I}^{\diff}_{s}$
  & measured, rendered, and coherent-difference world-frame images \\
$\C(\cdot;\vect{p}^{(t)}_i)$
  & GeoCrop, centered on the current twin state of building $i$ \\
$\vect{C}^{\real}_{i,s},\ \vect{C}^{\twin}_{i,s},\ \vect{C}^{\diff}_{i,s}$
  & paired GeoCrops for building $i$ from node $s$ \\
$g_i\in[0,1],\ \vect{r}_i\in\mathbb{R}^{6}$
  & movement gate and raw residual predicted for building $i$ \\
$\dphat{i}=g_i\,\vect{r}_i$
  & correction actually applied to the twin \\
$K$
  & number of backprojection focus heights \\
$S,\ N$
  & number of sensing nodes and of buildings in a scene \\
\bottomrule
\end{tabular}
\end{table}

\paragraph{The two CSI branches.}
For sensing node $s$ with known configuration $\vect{\xi}_s$, the physical observation is the measured complex channel state information
\begin{equation}
    \vect{H}^{\real}_{s}
    \in
    \mathbb{C}^{N_{\rm sym}\times N_{\rm tx}\times N_{\rm rx}\times N_{\rm sc}},
    \label{eq:csi_real}
\end{equation}
where $N_{\rm sym}$ is the number of sensing OFDM symbols or channel snapshots, $N_{\rm tx}$ and $N_{\rm rx}$ are the numbers of transmit and receive RF antenna elements, and $N_{\rm sc}$ is the number of observed orthogonal frequency subbands, a.k.a., subcarriers. The entry $H^{\real}_{s}[q,m,n,k]$ is the complex coefficient measured at sensing symbol $q$, between transmit element $m$ and receive element $n$, on subcarrier $k$; for a single snapshot $N_{\rm sym}=1$, the tensor reduces to $N_{\rm tx}\times N_{\rm rx}\times N_{\rm sc}$. Given the current twin $\vect{P}^{(t)}$, the same complex channel state information acquisition is synthesized with Sionna RT~\citep{hoydis2023sionna},
\begin{equation}
    \vect{H}^{\twin}_{s}(\vect{P}^{(t)})
    = \R\!\left(\vect{P}^{(t)};\vect{\xi}_s\right),
    \qquad
    \vect{H}^{\twin}_{s}
    \in
    \mathbb{C}^{N_{\rm sym}\times N_{\rm tx}\times N_{\rm rx}\times N_{\rm sc}} .
    \label{eq:csi_twin}
\end{equation}
Thus $\vect{H}^{\real}_{s}$ is what the physical environment produces and $\vect{H}^{\twin}_{s}$ is what the current twin predicts under the same radio acquisition. Matching the acquisition means that node pose, waveform, bandwidth, and array geometry are held fixed between the two branches, making the measured and rendered RF responses comparable. In simulation, both branches are generated using the same ray-tracing model;
their mismatch therefore arises from the geometric error, together with receiver
noise injected only into the physical branch.


\paragraph{Backprojection operator.}
All experiments use a single snapshot per acquisition ($N_{\rm sym}=1$), so the symbol dimension is dropped below. Let the world grid contain $N_x\times N_y$ horizontal locations and let $z_1,\ldots,z_K$ be the focus heights. For each height, coherent backprojection is the map
\begin{equation}
    \B_z:\ \mathbb{C}^{N_{\rm tx}\times N_{\rm rx}\times N_{\rm sc}}
    \rightarrow \mathbb{C}^{N_x\times N_y},
    \qquad
    \vect{I}^{\real}_{s,z} = \B_z(\vect{H}^{\real}_{s}),
    \quad
    \vect{I}^{\twin}_{s,z} = \B_z(\vect{H}^{\twin}_{s}),
    \label{eq:bp_operator}
\end{equation}
evaluated pointwise by Eq.~\eqref{eq:bp} of the main text. Two properties of $\B_z$ are used later. First, it is applied with identical parameters to both branches, so it introduces no asymmetry between measurement and hypothesis. Second, one world grid serves every sensing node, so pixel $(u,v)$ in any $\vect{I}_{s,z}$ corresponds to the same physical location $(x_u,y_v,z)$ regardless of which node observed it. Stacking the $K$ planes gives $\vect{I}^{\real}_{s},\vect{I}^{\twin}_{s}\in\mathbb{C}^{K\times N_x\times N_y}$, so vertical structure is retained as a channel axis rather than collapsed.

\paragraph{GeoCrop operator.}
For each building $i$, the current twin provides its estimated ground position $(x_i^{(t)},y_i^{(t)})$. We use this position to locate the building center on the world-frame backprojection grid and extract a fixed $H_p\times W_p$ window around it. The same window, centered at the same twin-provided location, is taken from the measured, rendered, and difference backprojections:
\begin{equation}
    \vect{C}^{b}_{i,s}
    = \C\!\left(\vect{I}^{b}_{s};\vect{p}^{(t)}_i\right)
    \in\mathbb{C}^{K\times H_p\times W_p},
    \qquad b\in\{\real,\twin,\diff\}.
    \label{eq:geocrop_op}
\end{equation}
The true building position $\vect{p}^{\star}_i$ is never used to define the crop. Consequently, if the current twin places the building incorrectly, the measured and rendered RF structures appear misaligned within the same local window. GeoCrop therefore presents the corrector with a local comparison around the building location proposed by the current twin, rather than the full backprojected scene. Crop centers lie on the discrete backprojection grid; the small fractional offset between the continuous twin position and the selected grid point is handled analytically as described in App.~\ref{app:arch}.

\paragraph{Synthetic branch and noise injection.}


For synthetic data, the branch corresponding to the measured RF observation is generated by rendering the true scene under the same sensing configuration, $\widetilde{\vect{H}}^{\real}_{s}=\R(\vect{P}^{\star};\vect{\xi}_s)$, and receiver noise is injected into the resulting physical-side backprojection. With $\vect{N}_{s,z}(\rho)\in\mathbb{C}^{N_x\times N_y}$ the noise field scaled to a target signal-to-noise ratio (SNR) $\rho$, the training pair is
\begin{equation}
    \vect{I}^{\real}_{s,z}
    = \B_z(\widetilde{\vect{H}}^{\real}_{s}) + \vect{N}_{s,z}(\rho),
    \qquad
    \vect{I}^{\twin}_{s,z}
    = \B_z\!\left(\R(\vect{P}^{(t)};\vect{\xi}_s)\right).
    \label{eq:noise_injection}
\end{equation}
Only the physical-side branch carries $\vect{P}^{\star}$ and only that branch receives noise; the twin branch is rendered noiselessly from $\vect{P}^{(t)}$. Both are produced by the same renderer, so in simulation the two branches differ by geometry and by the injected noise.

\paragraph{Measured branch.}
For measured data, $\vect{H}^{\real}_{s}$ is acquired directly from hardware. It already contains receiver noise, hardware impairments, and whatever the physical site contributes, so no synthetic noise is added. The twin branch is still rendered from the current twin and under the acquisition configuration of that measurement.


\paragraph{Supervision target and the gate label.}
In simulation, the initial twin $\vect{P}^{(0)}$ is generated from $\vect{P}^{\star}$ by a known corruption $\vect{\varepsilon}_i$ applied to a subset of buildings (App.~\ref{app:data}), so the regression target is available in closed form as $\Delta\vect{p}_i^{(0)}=-\vect{\varepsilon}_i$. The gate target $g_i^{\star}$ is derived from the 3D spatial displacement between $\vect{p}^{(t)}_i$ and $\vect{p}^{\star}_i$ using a small tolerance, so a building left exact by the generator is labeled as not requiring an update. 


\paragraph{Correspondence assumption.}
The residual is defined per building and therefore presumes a correspondence between an element of $\vect{P}^{(t)}$ and an element of $\vect{P}^{\star}$. A structure present in one and absent from the other therefore falls outside the formulation (Sec.~\ref{sec:limitations}). The same parameterization applies to any other scene element
described by a fixed parameter vector; buildings are used here because position, height, footprint, and orientation are the attributes for which public map products are documented to be unreliable (Sec.~\ref{sec:intro}).


\section{Representative Scene, World-Frame Backprojection, and GeoCrop}
\label{app:scenebp}

\begin{figure}[!t]
    \centering

    \begin{subfigure}[t]{0.45\textwidth}
        \centering
        \includegraphics[width=\linewidth]{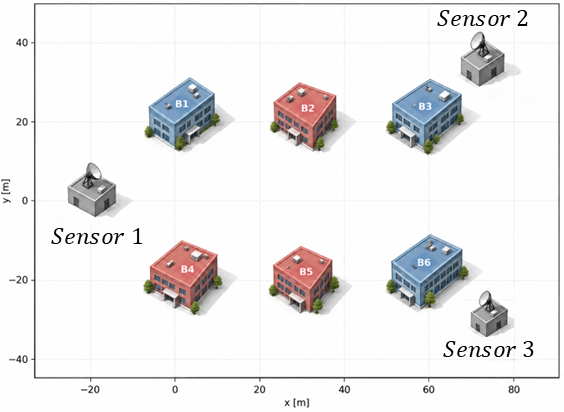}
        \caption{Representative six-building scene.}
    \end{subfigure}\hfill
    \begin{subfigure}[t]{0.45\textwidth}
        \centering
        \includegraphics[width=\linewidth]{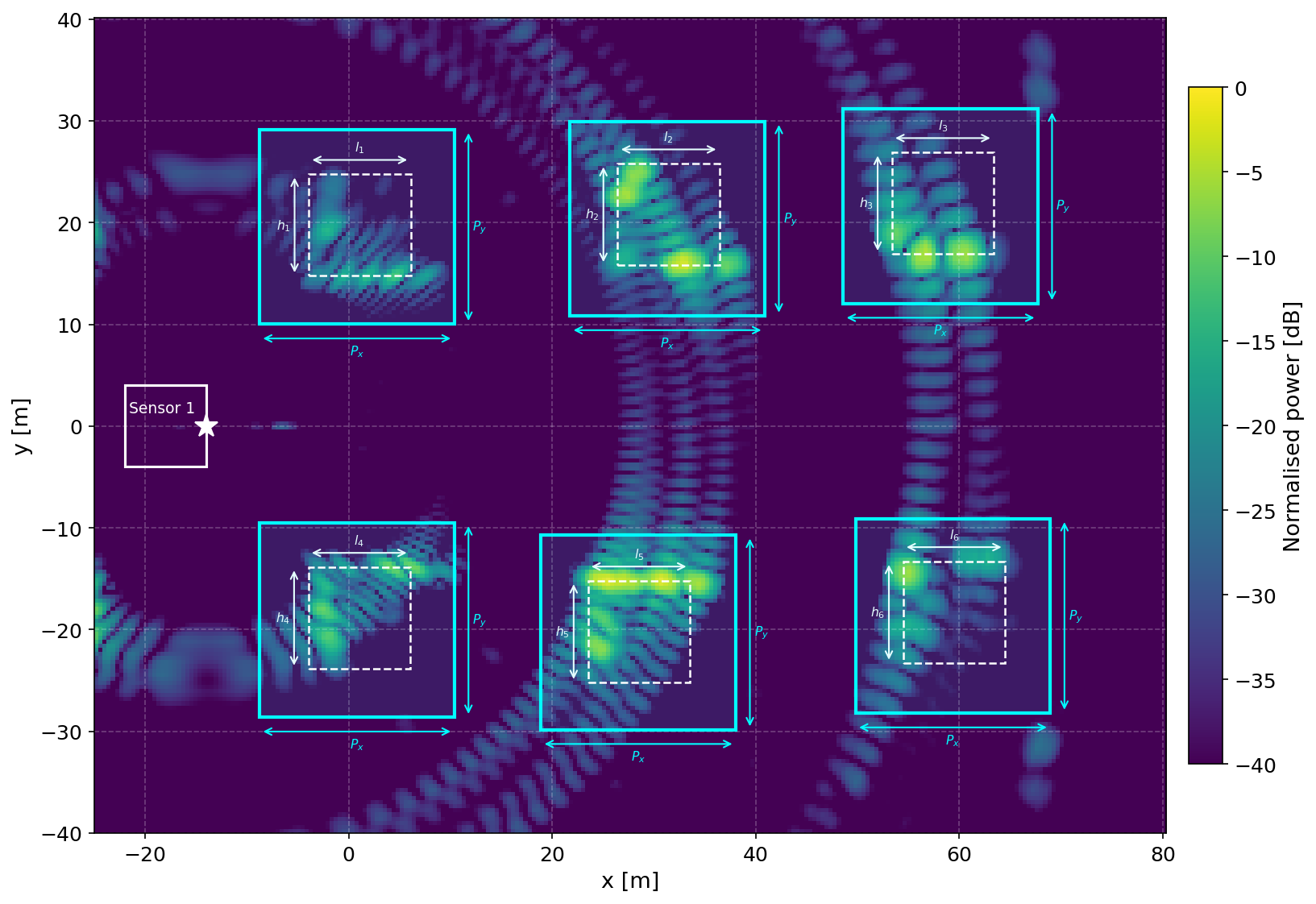}
        \caption{Sensor~1 backprojection.}
    \end{subfigure}

    \vspace{0.8em}

    \begin{subfigure}[t]{0.45\textwidth}
        \centering
        \includegraphics[width=\linewidth]{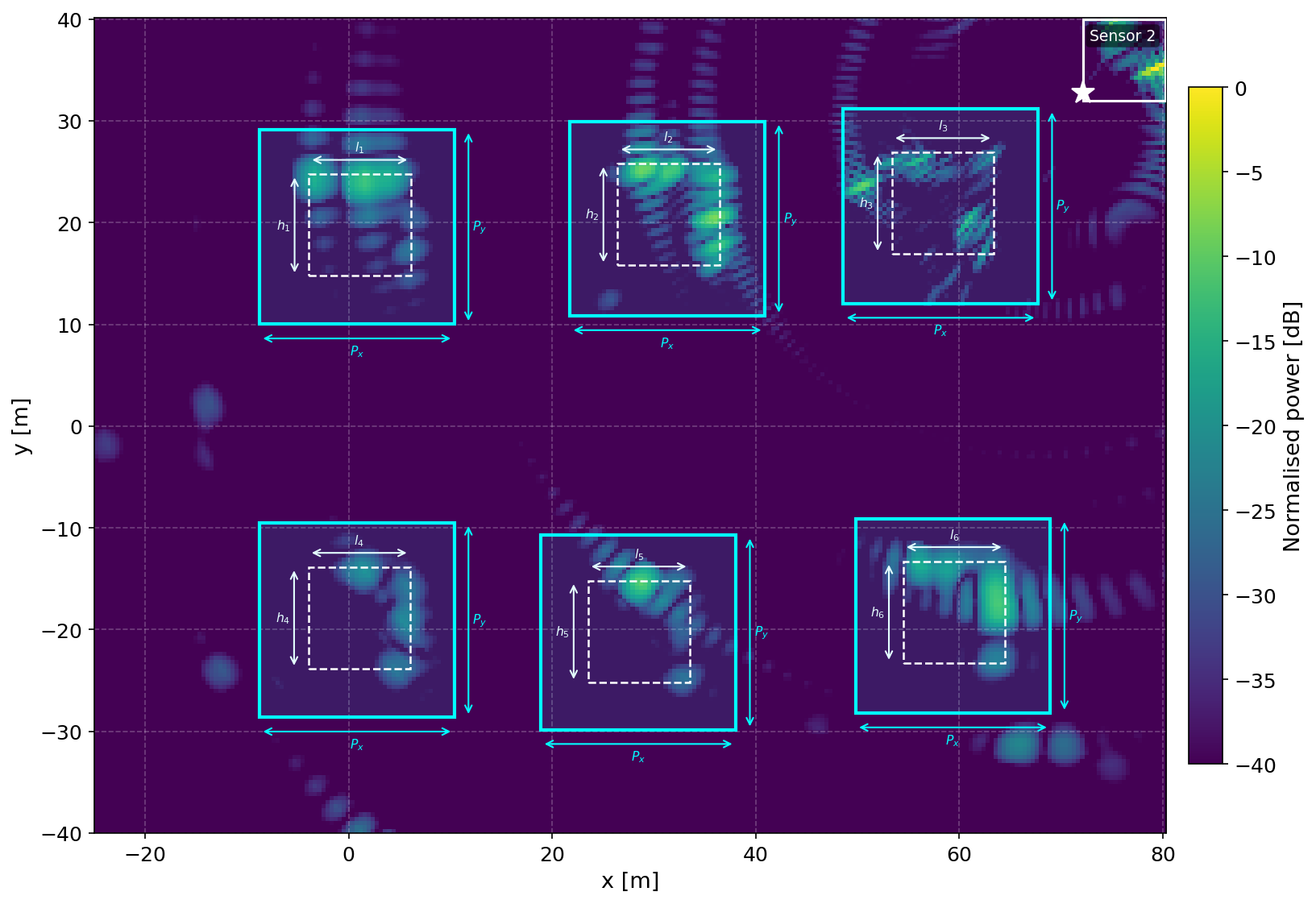}
        \caption{Sensor~2 backprojection.}
    \end{subfigure}\hfill
    \begin{subfigure}[t]{0.45\textwidth}
        \centering
        \includegraphics[width=\linewidth]{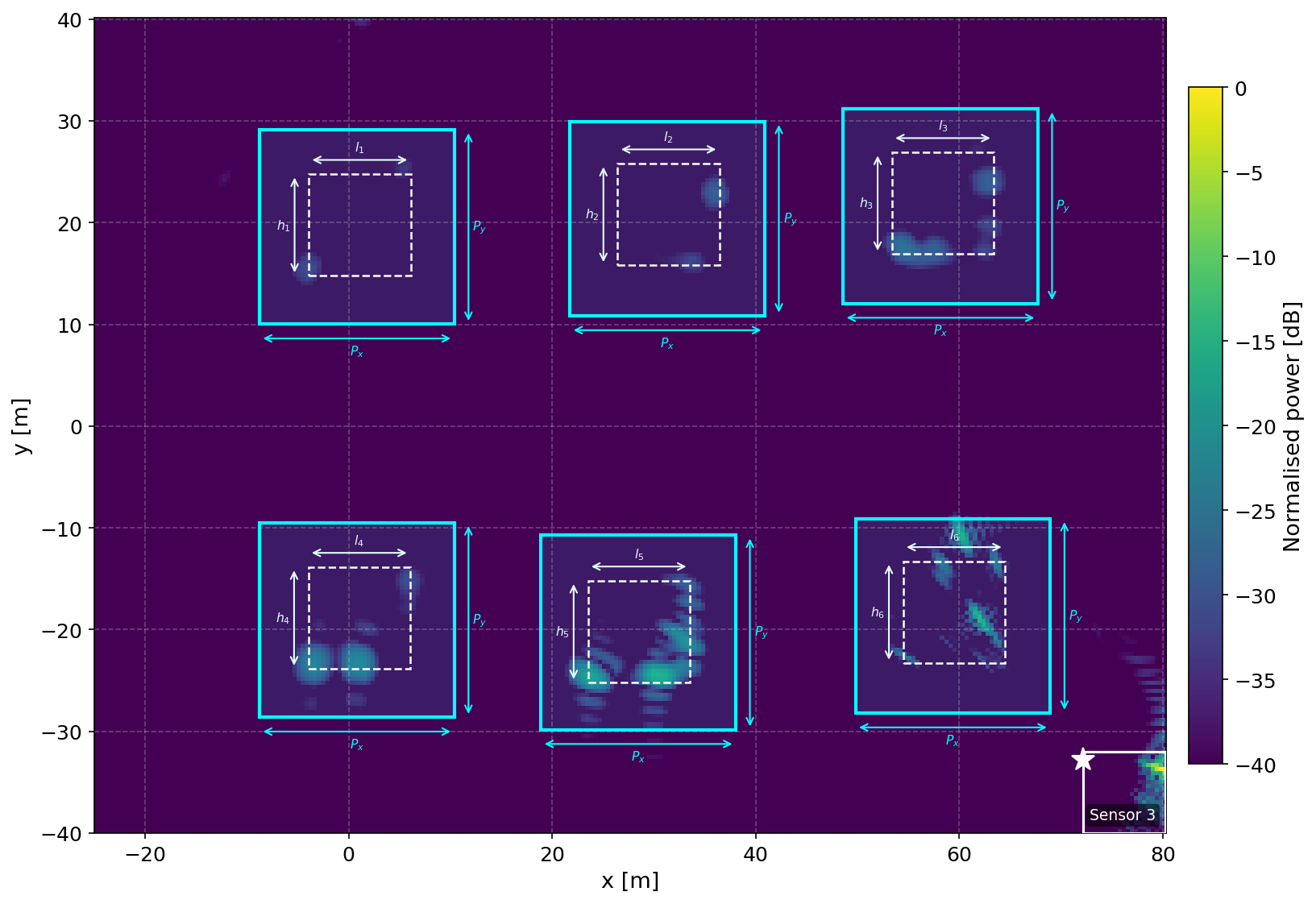}
        \caption{Sensor~3 backprojection.}
    \end{subfigure}

    \caption{\textbf{Representative scene and complementary world-frame backprojections at \(z = 9 m\).}
    (a) A six-building synthetic scene with three separated rooftop sensing nodes. (b)--(d) Corrupted-twin backprojections from Sensing node~1--3 on the same metric world grid. Dashed boxes indicate the current-twin footprint
    priors and solid cyan boxes indicate the fixed GeoCrop windows, which are applied identically to the corresponding measured backprojections. The RF response of a building depends strongly on sensing viewpoint:
    structures that are prominent from one node can be weak or poorly resolved from another, motivating per-node processing followed by cross-sensor fusion.}
    \label{fig:scene_bp}
\end{figure}

This section illustrates what the learned corrector receives for each building. Figure~\ref{fig:scene_bp} shows a representative six-building scene together with world-frame backprojections from all three sensing nodes. Because all nodes are mapped onto the same metric $x$--$y$ grid, the same building occupies the same physical region in every backprojection, while the RF structure observed within that region can differ substantially with sensing viewpoint.

\paragraph{What the world frame buys.}
The backprojections in Fig.~\ref{fig:scene_bp}b--d are expressed in the same $x$--$y$ metric coordinates as the physical scene in Fig.~\ref{fig:scene_bp}a. Backprojected power concentrates around wall and corner responses visible to each sensing node, so the spatial RF evidence is already registered to the building geometry in the twin. More importantly, a GeoCrop centered on building $i$ selects the same physical region from every node, removing the need for the learned corrector to first determine which responses across different sensing nodes correspond to the same building.

\paragraph{Why multiple sensing viewpoints help.}
Although the coordinate frame is shared, the RF evidence is not identical across sensing nodes. Visibility, illumination, occlusion, and propagation geometry cause different walls and corners to dominate from different viewpoints. For example, the responses around Buildings~1--3 are substantially weaker and less clearly resolved from Sensor~3 than from Sensor~2, while Sensor~3 provides strong evidence for several buildings in the lower part of the scene. Thus, no single node is uniformly informative for every building. \trace{} therefore preserves the per-node GeoCrops separately and combines their evidence only after each view has been encoded, allowing a strong viewpoint to complement a weak or ambiguous one.

\paragraph{Why several focus heights.}
Backprojection at a single height $z$ focuses the scattering that is consistent with that plane and defocuses the rest, so a single plane under-determines vertical extent. Repeating the operation at
$z_1,\ldots,z_K$ and stacking the complex planes exposes how the response of a given footprint changes with height, which is the evidence available for the height component of the residual. The stack is carried through the encoder as input channels, so adding heights changes the input depth and not the structure of the model.

\paragraph{What the crop is centered on, and why it is never centered on truth.} The solid cyan rectangles are the GeoCrop windows. Their centers come from $\vect{p}^{(t)}_i$, the state the twin currently holds, and the identical windows are applied to the measured backprojection. Centering the measured crop on $\vect{p}^{\star}_i$ would place the true structure at the center of every window and would remove exactly the displacement the model is asked to estimate; it would also require the ground truth at inference time, which is unavailable in deployment. Using the current twin center instead keeps the operation applicable whenever the twin is available.

\paragraph{What a wrong twin looks like inside one window.}
Because the window is anchored to the twin, a geometric error appears as a relative misalignment between the two branches inside a shared local frame. A planar displacement shifts the measured wall response away from the rendered one; an incorrect footprint length or width changes the extent of the measured response relative to the rendered one; an incorrect yaw rotates the measured wall and corner structure with respect to the rendered structure. The window is fixed in metric size, so these comparisons are expressed in the same units for every building in every scene, which is what makes one shared encoder applicable to all of them.


\section{Learned Corrector: Layer-Level Definitions}
\label{app:arch}

Section~\ref{sec:corrector} presents the corrector as four design decisions. This appendix gives the algebra, in the order in which a scene passes through the model, and states for each stage what it is for, what it computes, and what follows from it. Batch dimensions are omitted throughout. A scene passes through ten stages: the paired GeoCrops together with the coherent residual stream; the shared complex encoder; sensor-pose FiLM; real-to-twin cross-attention; CueFusion; set-valued cross-sensor fusion; the current shape prior; relative-geometry building attention; the gate and residual heads; and the sub-pixel correction applied to the planar output. The subsections below follow that order, and the numerical settings for every stage are collected in Table~\ref{tab:architecture}.

\subsection{Paired inputs and the coherent residual stream}
\label{app:paired_inputs}
\paragraph{Role.}
For each building and sensing node, the corrector receives three aligned complex GeoCrops: the measured RF response, the response predicted by the current twin, and their explicit difference. The first two branches preserve the full local RF structure produced by the physical scene and by the current twin hypothesis, while the third branch makes their disagreement directly available to the network.

For every building $i\in\{1,\ldots,N\}$ and node
$s\in\{1,\ldots,S\}$, GeoCrop provides
\[
\vect{C}^{\real}_{i,s},
\qquad
\vect{C}^{\twin}_{i,s}
\in
\mathbb{C}^{K\times H_p\times W_p}.
\]
Before cropping, we also form the complex world-frame difference
\begin{equation}
    \vect{I}^{\diff}_{s}
    =
    \vect{I}^{\real}_{s}
    -
    \vect{I}^{\twin}_{s},
    \qquad
    \vect{C}^{\diff}_{i,s}
    =
    \C\!\left(
        \vect{I}^{\diff}_{s};
        \vect{p}^{(t)}_i
    \right),
    \label{eq:idiff}
\end{equation}
using exactly the same GeoCrop window as for the measured and rendered branches. Thus the learned corrector is given
\[
\left(
\vect{C}^{\real}_{i,s},
\vect{C}^{\twin}_{i,s},
\vect{C}^{\diff}_{i,s}
\right)
\]
for each building-node pair.

\paragraph{Consequence.}
The difference branch is not an additional observation; it is an explicit residual computed from the measured and rendered RF before either is passed through the learned encoder. Local responses that are similar in the physical scene and the current twin tend to cancel, while disagreement between them is emphasized. At the same time, \trace{} retains the original measured and rendered branches rather than giving the network only their difference. This allows the corrector to use both the RF structures being compared and the residual between them when estimating the geometric update.

\subsection{Shared CV-CNN encoder}
\label{app:complex_encoder}
\paragraph{Role.} Turn a local complex patch into tokens without discarding phase, using one set of weights for every building, node, and branch.

A shared complex-valued CNN (CV-CNN) $\E_{\theta}$ acts as the local RF feature extractor and tokenizer:
\begin{equation}
    \vect{F}^{b}_{i,s} = \E_{\theta}\!\left(\vect{C}^{b}_{i,s}\right)
    \in \mathbb{R}^{T\times D_{\rm emb}},
    \qquad b\in\{\real,\twin,\diff\},
    \label{eq:complex_encoder}
\end{equation}
with $T$ tokens of dimension $D_{\rm emb}$. At the output of the final complex stage, the real and imaginary parts of each feature are concatenated, giving $2\times128=256$ real channels, which a linear projection maps to $D_{\rm emb}$; phase information therefore reaches the tokens without being reduced to magnitude. The $K$ focus heights enter as convolutional input channels, and a spatial positional encoding is added so that the location of a feature within the window is retained by the token sequence. Each branch therefore yields a tensor of size
$S\times N\times T\times D_{\rm emb}$ per scene.

\paragraph{Consequence.} Cardinality generalization is a property of weight sharing rather than of the attention layers that follow. This is why the same checkpoint accepts the four-, eight-, and ten-building scenes of Sec.~\ref{sec:results_invariance} without modification.

\subsection{Sensor-pose FiLM}
\label{app:film}
\paragraph{Role.} Tell the shared encoder from which viewpoint a patch was acquired, without giving any node its own parameters.

For node $s$ we form the pose vector
\begin{equation}
    \vect{\xi}^{\rm pose}_s =
    [\bar{x}_s,\bar{y}_s,\bar{z}_s,\cos\psi_s,\sin\psi_s,\cos\tau_s,\sin\tau_s]^{\top},
\end{equation}
where $(\bar{x}_s,\bar{y}_s,\bar{z}_s)$ is the normalized sensing-node position, $\psi_s$ the boresight azimuth, and $\tau_s$ the elevation tilt. A small network maps it to feature-wise modulation parameters
$(\vect{\gamma}_s,\vect{\beta}_s)$:
\begin{equation}
    \vect{M}^{b}_{i,s} = (1+\vect{\gamma}_s)\odot\vect{F}^{b}_{i,s}+\vect{\beta}_s,
    \qquad b\in\{\real,\twin,\diff\}.
    \label{eq:film}
\end{equation}
FiLM is initialized to the identity, so the encoder starts from the unconditioned solution and acquires viewpoint dependence during training.

\paragraph{Consequence.}
Sensor pose is provided as a conditioning value rather than encoded through a fixed sensor index. The same shared encoder can therefore process different numbers of sensing nodes and nodes observed from different positions or orientations without changing the network architecture. This enables the sensor-subset evaluations of Sec.~\ref{sec:results_invariance} and also allows the model to accept previously unseen sensing poses.

\subsection{Real-to-twin discrepancy attention}
\label{app:real_twin_attn}
\paragraph{Role.} Compare the measurement against the current hypothesis for one building from one node, before anything is mixed across nodes or buildings.

The $T$ measured tokens query the matching $T$ rendered tokens:
\begin{equation}
    \vect{Z}^{\probe}_{i,s}
    = \operatorname{MHA}\!\left(
        \vect{Q}=\vect{M}^{\real}_{i,s},\
        \vect{K}=\vect{M}^{\twin}_{i,s},\
        \vect{V}=\vect{M}^{\twin}_{i,s}\right)
    \in \mathbb{R}^{T\times D_{\rm emb}} .
    \label{eq:real_twin_attn}
\end{equation}

\paragraph{Consequence.} Placing the comparison first keeps it local in both senses: the query and the key come from the same building and the same viewpoint, so a mismatch found here cannot have been assembled from another building's evidence. It also gives the noisy branch a noiseless template to match against, since only the measured branch carries noise
(Eq.~\eqref{eq:noise_injection}); energy in the measured crop that corresponds to no rendered structure is attenuated rather than propagated. This asymmetry is one of the three properties the main text identifies as contributing to the low-SNR behavior of Sec.~\ref{sec:results_invariance}.

\subsection{CueFusion}
\label{app:cuefusion}
\paragraph{Role.} Preserve the raw evidence alongside the attention output, so that a small discrepancy is not compressed away by a single comparison route.

Writing $\vect{D}_{i,s}=\vect{M}^{\real}_{i,s}-\vect{M}^{\twin}_{i,s}$ for the feature-space difference,
\begin{equation}
    \vect{Z}^{\fuse}_{i,s} = \mathcal{P}_{\theta}\Big[
        \vect{Z}^{\probe}_{i,s} \,\Vert\,
        \vect{M}^{\real}_{i,s} \,\Vert\,
        \vect{M}^{\twin}_{i,s} \,\Vert\,
        \vect{D}_{i,s} \,\Vert\,
        \vect{M}^{\diff}_{i,s}\Big],
    \label{eq:cuefusion}
\end{equation}
with $\Vert$ concatenation and $\mathcal{P}_{\theta}$ a projection back to $D_{\rm emb}$. The last two cues are not duplicates of each other: $\vect{D}_{i,s}$ measures disagreement after nonlinear encoding, whereas $\vect{M}^{\diff}_{i,s}$ comes from subtracting the coherent complex backprojections first and encoding the residual afterwards. Since $\E_{\theta}$ is nonlinear,
\begin{equation}
    \E_{\theta}\!\left(\vect{C}^{\real}_{i,s}-\vect{C}^{\twin}_{i,s}\right)
    \neq
    \E_{\theta}\!\left(\vect{C}^{\real}_{i,s}\right)
    - \E_{\theta}\!\left(\vect{C}^{\twin}_{i,s}\right),
\end{equation}
and the two differences are genuinely different quantities.

\paragraph{Consequence.} The measured-to-rendered comparison is carried by three overlapping routes at this point: the cross-attention probe, the feature-space difference, and the coherent difference. The redundancy is deliberate, and App.~\ref{app:ablation_small} measures what it costs.

\subsection{Cross-sensor fusion}
\label{app:sensor_fusion}
\paragraph{Role.} Combine views of the same building from different nodes, allowing a well-placed view to compensate for a weak or ambiguous one.

For a fixed building $i$, the $T$ discrepancy tokens from all $S$ nodes are concatenated into one set and jointly attended, then pooled:
\begin{equation}
    \vect{Z}^{\fuse}_{i} = \operatorname{Concat}_{s=1}^{S}\vect{Z}^{\fuse}_{i,s}
    \in \mathbb{R}^{(ST)\times D_{\rm emb}},
    \qquad
    \widetilde{\vect{Z}}_{i} = \operatorname{SA}_{\rm sens}\!\left(\vect{Z}^{\fuse}_{i}\right),
    \label{eq:sensor_tokens}
\end{equation}
\begin{equation}
    \vect{Z}^{\agg}_{i} = \operatorname{MeanPool}\!\left(\widetilde{\vect{Z}}_{i}\right)
    \in \mathbb{R}^{D_{\rm emb}} .
    \label{eq:sensorfusion}
\end{equation}
No learned sensor-index embedding is introduced anywhere in
Eqs.~\eqref{eq:sensor_tokens} and~\eqref{eq:sensorfusion}. Random sensing-node subsets are presented during training so that reduced node sets are in distribution.

\paragraph{Consequence.} The self-attention determines how well the views are combined, which App.~\ref{app:ablation_fusion} measures. The absence of sensor slots determines that the output does not depend on the order in which nodes are presented, which App.~\ref{app:ablation_permutation} derives. The two properties are independent and are attributed separately.

\subsection{Current shape prior}
\label{app:shape_prior}
\paragraph{Role.} Tell the model what geometry the twin currently assumes for this building, without telling it where the building sits on the map.

The non-location components of the current twin prior are embedded by a small MLP $\phi_{\rm shape}$ and added to the sensing-aggregated token:
\begin{equation}
    \vect{s}_{i} = \phi_{\rm shape}\!\left(
        h_i^{(t)},\ \ell_i^{(t)},\ w_i^{(t)},\
        \sin 2\psi_i^{(t)},\ \cos 2\psi_i^{(t)}\right)
    \in \mathbb{R}^{D_{\rm emb}},
    \qquad
    \widetilde{\vect{Z}}^{\agg}_{i} = \vect{Z}^{\agg}_{i} + \vect{s}_{i} .
    \label{eq:shapeprior}
\end{equation}
The double-angle yaw encoding respects the $180^\circ$ symmetry of a rectangular footprint, so $\psi$ and $\psi+180^\circ$ are represented identically.

\paragraph{Consequence.} The rendered branch already encodes the current hypothesis in the RF domain; this path adds it in the parameter domain, which is the space the residual is predicted in. Absolute ground coordinates $(x_i^{(t)},y_i^{(t)})$ are excluded here, so this path cannot supply the building's map location.

\subsection{Relative-geometry building attention}
\label{app:building_attention}

\paragraph{Role.}After fusing the sensing-node observations for each building, \trace{} allows buildings to exchange information with other buildings in the same scene. This provides scene context that can help interpret ambiguous local RF evidence. The interaction is based on the relative geometry between buildings, rather than their absolute $(x,y)$ coordinates.

Every ordered building pair $(i,j)$ is described by relative geometry only:
\begin{equation}
\begin{aligned}
    \vect{R}_{ij} = \phi_{\rm geo}\big(
        &\Delta x_{ij},\ \Delta y_{ij},\ \Delta h_{ij},\ d_{ij},\
        \Delta\ell_{ij},\ \Delta w_{ij}, \\
        &\sin(2\Delta\psi_{ij}),\ \cos(2\Delta\psi_{ij})\big)
    \in \mathbb{R}^{D_{\rm emb}},
\end{aligned}
    \label{eq:relative_geometry}
\end{equation}
where $\Delta x_{ij}=x_i^{(t)}-x_j^{(t)}$ and analogously for the remaining parameters, and $d_{ij}$ is the 3D separation. The MLP $\phi_{\rm geo}$ is shared across all pairs. Attention across the $N$ buildings then adds the pairwise embedding to both key and value. With projections
$\vect{q}_{i}=\vect{W}_{Q}\widetilde{\vect{Z}}^{\agg}_{i}$,
$\vect{k}_{j}=\vect{W}_{K}\widetilde{\vect{Z}}^{\agg}_{j}$,
$\vect{v}_{j}=\vect{W}_{V}\widetilde{\vect{Z}}^{\agg}_{j}$,
\begin{equation}
\begin{aligned}
    e_{ij} &= \frac{\vect{q}_{i}^{\top}\left(\vect{k}_{j}+\vect{R}_{ij}\right)}{\sqrt{D_{\rm emb}}},
    \\
    \alpha_{ij} &= \operatorname{softmax}_{j}(e_{ij}),
    \\
    \vect{m}_{i} &= \sum_{j=1}^{N}\alpha_{ij}\left(\vect{v}_{j}+\vect{R}_{ij}\right),
\end{aligned}
    \label{eq:building_message}
\end{equation}
and the message is combined with the incoming token through a residual
connection, normalization, and a feed-forward block $G_{\theta}$:
\begin{equation}
    \vect{Z}^{\bld}_{i} = G_{\theta}\!\left(\widetilde{\vect{Z}}^{\agg}_{i}+\vect{m}_{i}\right)
    \in \mathbb{R}^{D_{\rm emb}} .
    \label{eq:building_token}
\end{equation}


\paragraph{Consequence.}
Because the building-attention stage uses only relative geometry between buildings Eq.~\eqref{eq:relative_geometry}, its representation depends on how buildings are arranged with respect to one another rather than on their absolute location in the scene.
For example, translating all buildings by the same amount leaves their pairwise relative geometry unchanged. This reduces the opportunity for the scene-reasoning module to memorize location-specific patterns from the training layouts and supports generalization to unseen building arrangements. This is what the OOD-Layout evaluations of Sec.~\ref{sec:results_representation} and App.~\ref{app:closedloop} test. 

\subsection{Gated prediction heads and sub-pixel correction}
\label{app:prediction_heads}
\paragraph{Role.} Decide separately whether a building should be changed and
how it should be changed.

A movement head and a regression head produce
\begin{equation}
\begin{aligned}
    a_i &= f_{\rm gate}\!\left(\vect{Z}^{\bld}_{i}\right),
    \qquad g_i = \sigma(a_i)\in[0,1], \\
    \vect{r}_{i} &= f_{\rm res}\!\left(\vect{Z}^{\bld}_{i}\right)
    = [\Delta x_i,\Delta y_i,\Delta h_i,\Delta\ell_i,\Delta w_i,\Delta\psi_i]^{\top},
\end{aligned}
\end{equation}
and the applied correction is $\dphat{i}=g_i\,\vect{r}_{i}$, as in Eq.~\eqref{eq:loop}. Both heads read the same token, so the decision to act and the estimate of the action are conditioned on identical evidence. At evaluation the gate is used in its soft form.

GeoCrop centers lie on the discrete backprojection grid, so the continuous twin location can sit slightly away from the nearest pixel used as the crop center. That fractional $x$--$y$ offset is recorded deterministically and removed analytically from the predicted planar residual before the gate is applied. It is not supplied as a learned input, which keeps grid quantization from being absorbed into the geometric estimate. The full-scene baselines crop nothing and therefore require no such correction.

\paragraph{Consequence.}
Because some buildings in a scene are already correct while others contain geometric errors, the gate allows \trace{} to apply a correction only where one is needed and to suppress unnecessary updates to buildings that are already well aligned. Its contribution is evaluated in
App.~\ref{app:ablation_gate}.



\section{Preserving the Complex RF Response}
\label{app:complex}

\begin{table}[h]
\centering
\caption{\textbf{Complex versus amplitude-only backprojection.} 3D position RMSE on the scene-disjoint test split with everything else in the pipeline held fixed; lower is better. The uncorrected twin is \SI{2.202}{m}.}
\label{tab:complex_vs_amp}
\small
\setlength{\tabcolsep}{8pt}
\begin{tabular}{lc}
\toprule
\textbf{Input} & \textbf{3D RMSE [m]} $\downarrow$ \\
\midrule
Amplitude only (\trace{})& 0.350 \\
Complex (\trace{}) & \textbf{0.302} \\
\bottomrule
\end{tabular}
\end{table}

Section~\ref{sec:frontend} states that the backprojection is complex and that phase is useful for geometric displacement. This appendix reports the experiment behind that statement and explains the physical reason.


As shown in Table~\ref{tab:complex_vs_amp}, replacing the complex-valued backprojection with an amplitude-only input, with everything else in the pipeline held fixed, increases 3D position RMSE from \SI{0.302}{m} to \SI{0.350}{m}. Retaining the full complex response therefore reduces the remaining error by 13.7\%.

The reason is a difference in how the two quantities respond to a small displacement. Backprojected magnitude at a world point reflects how much energy is consistent with that point, and it changes appreciably only when the displacement is large enough to move scattering structure between resolution cells. The phase of the same quantity is set by the round-trip path length in units of the carrier wavelength, which at \SI{28}{GHz} is about
\SI{1.07}{cm}. Moving a scatterer by a fraction of a wavelength can therefore leave $|\vect{I}|$ nearly unchanged while rotating $\arg(\vect{I})$ substantially. Phase supplies sensitivity in exactly the regime where magnitude is least informative, which is the regime the corrector operates in once the twin is already approximately right.



\section{Effect of the Number of Backprojection Focus Heights}
\label{app:height_ablation}

The world-frame representation uses six backprojection focus heights, $z\in\{0,3,6,9,12,15\}$\,m. As discussed in Sec.~\ref{sec:frontend}, backprojection at different elevations exposes how the RF response changes with vertical position, providing information that is not available from a single horizontal slice. We quantify how strongly the trained models use this multi-height representation.

All models are trained using all six focus planes. At test time, we progressively reduce the number of available planes while keeping the trained model completely unchanged. Unavailable planes are set to zero. Thus, the network architecture, tensor dimensions, and learned weights remain fixed; only the number of available backprojection heights changes.

Because six planes give $2^6-1=63$ possible non-empty subsets, we evaluate all 63 subsets for each model. For a given number of available heights, we report the average performance across all subsets of that size. This allows us to ask directly how correction accuracy changes as the model is given progressively more vertical RF information.


\begin{figure}[t]
    \centering
    \includegraphics[width=0.72\linewidth]{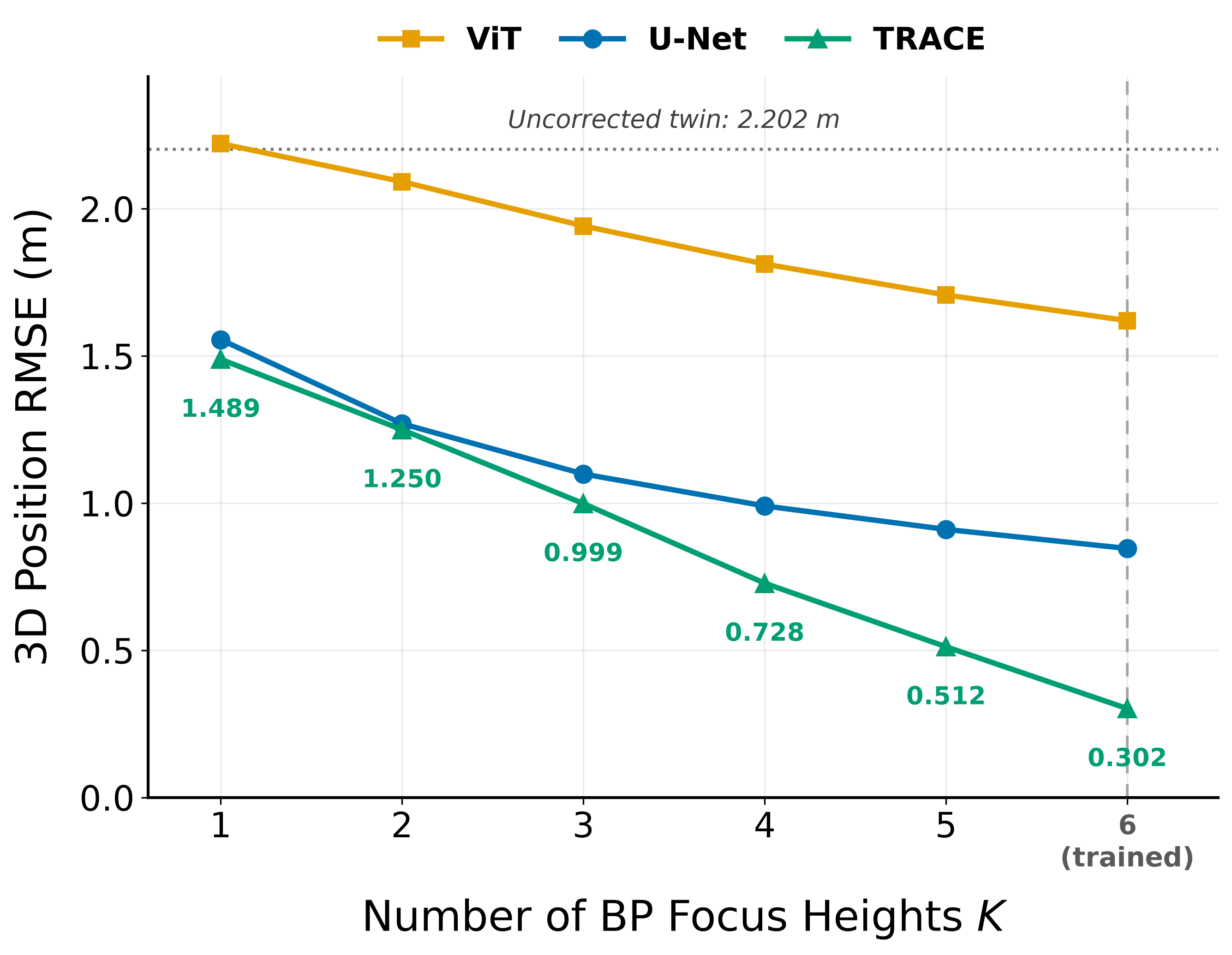}
    \caption{\textbf{Correction accuracy improves as more backprojection focus heights are available.}
    Mean 3D position RMSE as the number of available focus heights increases. Each point averages over all subsets of the corresponding size. All models
    are trained with six heights and evaluated with frozen weights.}
    \label{fig:height_ablation}
\end{figure}


\begin{table}[t]
\centering
\caption{\textbf{Effect of the number of backprojection focus heights $K$ on \trace{}.} The model is trained with all six focus heights; at test time, only $K$ heights are available and the rest are set to zero. Results are averaged over all subsets of each size. Lower is better.}
\label{tab:height_ablation}
\small
\setlength{\tabcolsep}{9pt}
\begin{tabular}{lccc}
\toprule
\textbf{Method} & \textbf{2D [m]} $\downarrow$ & \textbf{Height [m]} $\downarrow$ & \textbf{3D [m]} $\downarrow$ \\
\midrule
Uncorrected twin & 1.878 & 1.151 & 2.202 \\
\midrule
\trace{}, $K=1$ & 0.720 & 1.268 & 1.489 \\
\trace{}, $K=2$ & 0.430 & 1.157 & 1.250 \\
\trace{}, $K=3$ & 0.312 & 0.942 & 0.999 \\
\trace{}, $K=4$ & 0.255 & 0.675 & 0.728 \\
\trace{}, $K=5$ & 0.224 & 0.455 & 0.512 \\
\trace{}, $K=6$ (full) & \textbf{0.205} & \textbf{0.221} & \textbf{0.302} \\
\bottomrule
\end{tabular}
\end{table}

\paragraph{More focus heights consistently improve correction accuracy.}
Figure~\ref{fig:height_ablation} shows a clear trend: as more focus planes are made available, the geometric correction becomes more accurate. For \trace{},
mean 3D position RMSE decreases from \SI{1.489}{m} with one focus height to \SI{1.250}{m} with two, \SI{0.999}{m} with three, and ultimately \SI{0.302}{m} with all six. The U-Net and ViT show the same overall trend,
although their final errors remain substantially higher. This confirms that the different backprojection heights provide complementary information rather than
redundant copies of the same RF observation.

\paragraph{The largest benefit appears in building-height estimation.}
The effect is especially strong for the vertical component of the correction. With only one focus plane, \trace{} has a building-height RMSE of \SI{1.268}{m}, slightly worse than the \SI{1.151}{m} uncorrected twin.
As additional heights are restored, height RMSE decreases steadily to \SI{0.221}{m} with all six planes. Planar position also improves, from \SI{0.720}{m} with one height to \SI{0.205}{m} with six, but the larger change in height error shows why observing the scene at multiple focus elevations is particularly important for recovering vertical geometry.

Overall, the experiment shows that the six-plane backprojection stack is an important part of the RF representation: a single focus plane retains some
useful information, but progressively adding focus heights produces increasingly accurate geometric correction, with the strongest effect on building height.

\section{Comparison of \trace{} with Classical Baselines}
\label{app:classical_registration}

The main experiments compare \trace{} with the learned ViT and U-Net
baselines. We additionally ask whether, once the measured and twin-rendered RF
have been backprojected into the same world frame and localized with GeoCrop,
the remaining planar correction can be recovered by a standard deterministic
image-registration method. If so, the learned corrector would provide limited
benefit beyond the representation itself.

We test this directly using two training-free registration baselines:
\emph{phase correlation} and \emph{normalized cross-correlation} (NCC).
Importantly, these methods are asked only to estimate the planar displacement
$(\Delta x,\Delta y)$ and are evaluated only on 2D position error; they are not
penalized for not recovering height, footprint, or yaw.

\paragraph{Classical registration baselines.}
For each building and sensing node, we use the same building-centered spatial
region provided by GeoCrop. To obtain a conventional registration image, the
complex multi-height RF response is collapsed to a real-valued power image,
\[
P(\vect{x}) = \sum_{z} |C_z(\vect{x})|^2 .
\]
Phase correlation computes the normalized cross-power spectrum between the
measured and twin-rendered power images and obtains the planar displacement from
the peak of its inverse Fourier transform. NCC instead searches for the spatial
offset that maximizes normalized similarity between the two images. Each method
produces one displacement estimate per sensing node, and the node-wise estimates
are combined using a confidence-weighted average of their correlation peaks.
Neither baseline contains learned parameters.

We evaluate both methods on the same 5{,}400 scene-disjoint test samples, using the same corrupted twins
and \SI{15}{dB} sensing condition as the learned models. The ViT and U-Net results from the main evaluation are included for reference.

\paragraph{Why not optimize the twin through the renderer?}
A per-scene alternative is to search for building parameters that minimize the measured-to-rendered discrepancy, either by gradient descent through a differentiable ray tracer or by derivative-free search. Both face two obstacles in this setting. First, meter-scale building errors change which paths exist, yet ray-tracing gradients do not account for paths that appear or disappear. Second, at \SI{28}{GHz} the phase of each path wraps every $\lambda/2\approx\SI{0.5}{cm}$ of path change, so a coherent discrepancy is highly non-convex in building position over the meter-scale errors considered here. Search-based optimization avoids gradients but requires many ray-tracing calls per building and per scene. \trace{} instead amortizes this search into a single forward pass per correction step, reusable across scenes without per-deployment optimization.


\begin{table}[t]
\centering
\caption{\textbf{Classical registration baselines for planar correction.}
2D position RMSE on the 5{,}400-sample scene-disjoint test split at
\SI{15}{dB}. Phase correlation and NCC are training-free and operate on the
same building-centered spatial regions used by \trace{}. Error removed is
relative to the uncorrected twin.}
\label{tab:classical_baseline}
\small
\setlength{\tabcolsep}{8pt}
\begin{tabular}{llcc}
\toprule
Method & Type & 2D RMSE [m] $\downarrow$ & Error removed [\%] $\uparrow$ \\
\midrule
Corrupted twin & no correction & 1.878 & -- \\
\midrule
Phase correlation & training-free & 1.925 & $-2.5$ \\
NCC & training-free & 1.802 & $4.0$ \\
\midrule
ViT & learned & 1.483 & $21.0$ \\
U-Net & learned & 0.553 & $70.6$ \\
\textbf{\trace{}} & learned & \textbf{0.206} & $\mathbf{89.0}$ \\
\bottomrule
\end{tabular}
\end{table}

Table~\ref{tab:classical_baseline} shows that making the RF comparison local
does not reduce the problem to conventional image registration. Phase
correlation slightly increases the 2D position error, from \SI{1.878}{m} to
\SI{1.925}{m}, while NCC removes only 4.0\% of the initial error. In contrast,
the learned ViT and U-Net baselines recover substantially more of the displacement,
and \trace{} reduces the error to \SI{0.206}{m}, removing 89.0\%.

The classical methods struggle because the geometric displacement is not expressed as a uniform translation of the entire RF crop. Although the desired output here is only $(\Delta x,\Delta y)$, a GeoCrop contains RF energy from the target building together with multipath and surrounding structures that do not undergo the same spatial shift. The correlation peak can therefore be
dominated by RF structure that remains approximately stationary rather than by the displacement of the building being corrected. This effect is visible in the estimates: among buildings that are actually displaced, the median required planar correction is \SI{3.16}{m}, whereas phase correlation returns a median shift of only \SI{0.20}{m}.

This failure is not explained only by the presence of buildings that require no correction. Of the 32{,}400 building instances, 10{,}783 are displaced. Restricting evaluation to these buildings, phase correlation changes the 2D position RMSE from \SI{3.255}{m} before correction to \SI{3.323}{m} after correction. Thus, even when every evaluated building requires a nonzero planar update, direct registration does not recover the displacement reliably.

The comparison also highlights information used by the learned corrector that is absent from these classical baselines. First, the registration methods operate on collapsed power images and therefore do not use the complex RF phase available to \trace{}; App.~\ref{app:complex} shows that preserving the complex
response improves geometric correction. Second, the reliability of a displacement cue varies across sensing viewpoints. The classical methods fuse per-node shifts using a fixed confidence rule, whereas \trace{} preserves the node-specific representations and learns how to combine complementary views for each building. Finally, \trace{} can suppress unnecessary updates through its movement gate, while a registration method always returns an estimated shift. The moved-only experiment above shows, however, that the gate alone cannot explain the performance gap.


\section{Component Ablations and Model Design}
\label{app:ablation}

The main text establishes that the complete \trace{} system outperforms the full-scene baselines (Sec.~\ref{sec:results_representation}) and that the representation front end is what makes the correction rule transfer across layouts (Table~\ref{tab:repr}). Neither result says which parts of the \emph{learned corrector} contribute to the accuracy. This appendix asks that question directly, by removing one element of $f_\theta$ at a time. World-frame backprojection and GeoCrop are kept unchanged in these experiments so that each ablation isolates a component of the learned corrector; complementary experiments that vary the RF representation or the evidence available to the model are reported in Table~\ref{tab:repr}, Sec.~\ref{sec:results_falsification}, and Apps.~\ref{app:complex}, \ref{app:height_ablation}, and \ref{app:falsification}.


\paragraph{Protocol.}
Each row of Table~\ref{tab:ablation} drops a single design decision and retrains from scratch. Every variant shares the scene-disjoint split and its 5,400 held-out samples, the \SI{15}{dB} training SNR, the objective of Eq.~\eqref{eq:loss}, and the optimizer and schedule of
Table~\ref{tab:training}. 

\begin{table}[t]
\centering
\caption{\textbf{Effect of removing individual learned-corrector components.} Each row removes one component of the learned corrector and retrains the resulting model from scratch.
Par.\ [M] is the number of trainable parameters, in millions.
RMSE is the resulting 3D position error in meters.
Cost is the absolute increase in RMSE relative to full \trace{}, while Infl.\ [\%] expresses the same increase as a percentage of the full-\trace{} RMSE.
Removed [\%] is the fraction of the initial 3D position error of the corrupted twin that is eliminated after correction.
}
\label{tab:ablation}
\footnotesize
\setlength{\tabcolsep}{3pt}
\begin{tabular}{llccccc}
\toprule
Component removed & Replaced by & Par.\ [M] & RMSE & Cost & Infl.\ [\%] & Removed [\%] \\
\midrule
\emph{uncorrected twin} & \emph{no correction} & -- & 2.202 & -- & -- & 0.0 \\
\emph{none} & \textbf{full \trace{}} & 3.98 & \textbf{0.302} & -- & -- & \textbf{86.2} \\
\midrule
Cross-sensor attention & mean pooling only & 2.93 & 0.375 & $+0.072$ & $+23.8$ & 83.0 \\
Movement gate & raw residual applied & 3.98 & 0.332 & $+0.029$ & $+9.6$ & 84.9 \\
Sensor-pose FiLM & unconditioned encoder & 3.85 & 0.317 & $+0.014$ & $+4.6$ & 85.6 \\
Coherent residual stream & four-cue CueFusion & 3.91 & 0.315 & $+0.012$ & $+4.1$ & 85.7 \\
Real-to-twin attention & probe slot zeroed & 3.72 & 0.310 & $+0.007$ & $+2.4$ & 85.9 \\
\bottomrule
\end{tabular}
\end{table}

Every removal increases the error, so no component is inert, and the degradation is gradual: without its most costly component \trace{} still reaches \SI{0.375}{m} and removes 83.0\% of the initial 3D error, against 61.6\% for the U-Net and 26.4\% for the ViT under the same conditions (Sec.~\ref{sec:results_invariance}).

\subsection{Cross-sensor fusion}
\label{app:ablation_fusion}

This variant keeps the pooling of Eq.~\eqref{eq:sensorfusion} and removes the self-attention of Eq.~\eqref{eq:sensor_tokens}, so the sensing-node features are averaged rather than adaptively fused. Removing cross-sensor attention increases 3D position RMSE from \SI{0.302}{m} to \SI{0.375}{m}, a 23.8\% increase in error, making it the largest degradation among the component ablations.

The ablated model also contains fewer parameters than the full \trace{} (2.93\,M versus 3.98\,M), so part of the degradation may be attributable to reduced model capacity. Nevertheless, the performance gap becomes larger as the sensing conditions become more difficult. Removing cross-sensor attention increases 3D position RMSE by \SI{0.071}{m} at \SI{20}{dB},
\SI{0.120}{m} at \SI{5}{dB}, and \SI{0.373}{m} at
$-\SI{5}{dB}$. This trend is consistent with adaptive fusion becoming more important when the individual sensing views are noisy or ambiguous. Because the ablation also changes the parameter count, however, these results do not fully isolate fusion from model capacity.

\subsection{Movement gate}
\label{app:ablation_gate}

This variant applies the raw residual $\vect{r}_i$ directly instead of the gated correction $g_i\vect{r}_i$. The movement head and its BCE term in Eq.~\eqref{eq:loss} are retained, so the parameter count is unchanged at \(3.98 M\) and the training signal is unchanged. The comparison therefore isolates the use of the gate at inference rather than the presence of the gate in the model. Removing the movement gate increases 3D position RMSE from \SI{0.303}{m} to \SI{0.332}{m}, a 9.6\% increase in error. The gate helps by suppressing unnecessary updates to buildings that are already correctly aligned while allowing corrections for buildings with geometric errors.


\subsection{Sensor-pose FiLM and complementary comparison paths}
\label{app:ablation_small}

The remaining three rows remove, respectively, the sensor-pose conditioning of Eq.~\eqref{eq:film}, the coherent residual stream of
Eq.~\eqref{eq:idiff}, and the real-to-twin cross-attention of
Eq.~\eqref{eq:real_twin_attn}. Their 3D position RMSE increases are \SI{0.014}{m}, \SI{0.012}{m}, and \SI{0.007}{m}, respectively.

Removing sensor-pose FiLM increases 3D position RMSE from
\SI{0.303}{m} to \SI{0.317}{m}, a 4.6\% increase in error. FiLM has a different role from the remaining two ablations: it conditions the shared encoder on the position and orientation of the sensing node that produced each GeoCrop. Thus, the same encoder can process observations collected from different sensing geometries without introducing a separate encoder or sensor-specific parameter set for every node. Its modest in-distribution ablation cost should therefore not be interpreted as its only purpose; the module is primarily intended to make the RF representation explicitly aware of sensing viewpoint and to support applying the same model when the sensing geometry changes.

The coherent residual stream and real-to-twin attention instead provide two explicit ways to compare the measured and rendered RF responses. Removing them increases 3D position RMSE from \SI{0.303}{m} to \SI{0.315}{m} and \SI{0.310}{m}, respectively. These individual increases are small because CueFusion retains multiple overlapping comparison paths: real-to-twin cross-attention, the learned feature difference
$\vect{M}^{\real}_{i,s}-\vect{M}^{\twin}_{i,s}$, and the coherent residual feature $\vect{M}^{\diff}_{i,s}$. Removing one path therefore leaves the others available.

The real-to-twin attention ablation should be distinguished from the falsification experiment in which the twin-rendered RF is removed entirely. When real-to-twin attention is removed, both
$\vect{I}^{\real}$ and $\vect{I}^{\twin}$ are still available to the model, and their relationship can still be represented through the feature-difference and coherent-residual paths. This explains why removing only the cross-attention probe causes a relatively small increase in error.

\subsection{Replacing the attention-based corrector with a U-Net}
\label{app:unet_corrector}

The ablations above remove one component at a time. We also replace the entire attention-based corrector with a convolutional one: a U-Net regressor that receives the same paired GeoCrops, shape prior, and sensor-pose FiLM, trained under the same protocol. Unlike the full-scene U-Net of Table~\ref{tab:main}, this variant uses the \trace{} front end, so the comparison isolates the corrector architecture from the representation.

\begin{table}[h]
\centering
\caption{\textbf{Same GeoCrop front end, different corrector.} Fraction of initial 3D position error removed across evaluation SNR; both models are trained at \SI{15}{dB}. Higher is better.}
\label{tab:unet_corrector}
\small
\setlength{\tabcolsep}{6pt}
\begin{tabular}{lcccccc}
\toprule
\textbf{Corrector} & \SI{-5}{dB} & \SI{0}{dB} & \SI{5}{dB} & \SI{10}{dB} & \SI{15}{dB} & \SI{20}{dB} \\
\midrule
U-Net on GeoCrops & 25.6 & 55.4 & 72.6 & 80.4 & 83.1 & 83.7 \\
\trace{} & \textbf{52.8} & \textbf{72.6} & \textbf{81.3} & \textbf{84.9} & \textbf{86.2} & \textbf{86.6} \\
\bottomrule
\end{tabular}
\end{table}

Table~\ref{tab:unet_corrector} shows two effects. At high SNR, the U-Net on GeoCrops already removes most of the error, confirming that the building-centered representation carries much of the gain. As SNR decreases, the gap widens: at \SI{-5}{dB}, \trace{} removes twice as much error (52.8\% versus 25.6\%). The attention-based comparison of measured and rendered crops, followed by adaptive fusion across nodes, is therefore what makes the correction substantially more robust when individual views are noisy.

\subsection{Generalization to Unseen Sensing Poses}
\label{app:pose_generalization}

Sensor-pose FiLM conditions each sensing-node representation on the node's position and orientation (App.~\ref{app:film}). We therefore test whether a corrector trained with the nominal sensing geometry can be applied directly to sensing poses not observed during training. All models use frozen \SI{15}{dB} checkpoints; no retraining, fine-tuning, or test-time adaptation is performed.

We perturb the sensing geometry through joint boresight-yaw rotations up to $\pm30^\circ$, elevation-tilt changes up to $\pm15^\circ$, and combined yaw--tilt offsets. We also include a more severe stress test in which two sensing nodes are displaced by \SI{38}{m} and rotated by $\pm51^\circ$. Across the 17 tested pose conditions, each condition is independently ray-traced with 700 samples. For the rotation
and tilt sweeps, scene layouts and twin corruptions are held fixed so that changes in performance are attributable to the sensing pose. 

\begin{table}[t]
\centering
\caption{\textbf{Generalization to unseen sensing poses.}
Minimum fraction of initial 3D position error removed across all tested pose
conditions using frozen \SI{15}{dB} checkpoints; higher is better.}
\label{tab:pose_robustness}
\small
\setlength{\tabcolsep}{10pt}
\begin{tabular}{lc}
\toprule
Model & Worst-case error removed [\%] $\uparrow$ \\
\midrule
\textbf{\trace{}} & \textbf{79.6} \\
U-Net & 54.8 \\
ViT & 18.1 \\
\bottomrule
\end{tabular}
\end{table}

Table~\ref{tab:pose_robustness} summarizes pose robustness by reporting the \emph{worst-case error removed}, defined as the smallest fraction of the initial 3D position error removed over all tested sensing-pose conditions. Thus, the column asks a simple question: under the most challenging pose perturbation we tested, how much corrective ability does each frozen model retain? For all three models, the minimum occurs in the \SI{38}{m} displacement stress test.
Even there, \trace{} removes 79.6\% of the initial error, compared with 54.8\% for the U-Net and 18.1\% for the ViT. The rotation and tilt sweeps show the same trend: \trace{} reaches only \SI{0.407}{m} RMSE in its worst rotation/tilt
condition, still below the U-Net's \SI{0.846}{m} nominal-pose error.
Together, these results show that the frozen \trace{} corrector remains effective under sensing geometries not observed during training, consistent with the role of sensor-pose FiLM conditioning in the learned corrector.

\subsection{Permutation invariance}
\label{app:ablation_permutation}

The sensing nodes are treated as an unordered set rather than assigned to fixed input positions. Therefore, changing the order in which the same sensing nodes are provided to the model does not change the prediction. In our test, the output is unchanged to three decimal places across all six permutations of the three sensing nodes.

This property follows from the architecture: all nodes are processed by the same shared encoder, sensor pose is provided through FiLM rather than a sensor-specific index, and the cross-sensor features are combined with permutation-invariant pooling. Thus, \trace{} does not depend on an arbitrary ordering such as Sensor~1, Sensor~2, Sensor~3. Cross-sensor attention determines how information from the different views is combined, while the set-based formulation makes the result independent of their input order.


\section{Additional Falsification Results}
\label{app:falsification}

Section~\ref{sec:results_falsification} shows that accurate correction requires both the measured and twin-rendered RF and that the two branches must correspond to the same scene. The main text reports these experiments using 3D position error. Here we provide the corresponding results for the remaining geometric parameters.

\paragraph{Input blinding across all geometric parameters.}
We use the same trained \trace{} model as in the main experiment and keep its weights frozen. To remove one RF branch, the corresponding world-frame backprojection is set to zero before GeoCrop. The current twin state $\vect{P}^{(t)}$ is unchanged. Thus, these tests measure how the trained paired-input model behaves when measured RF, twin-rendered RF, or both are unavailable; no model is retrained for the reduced inputs.

\begin{figure}[htbp]
    \centering
    \includegraphics[width=0.8\textwidth]{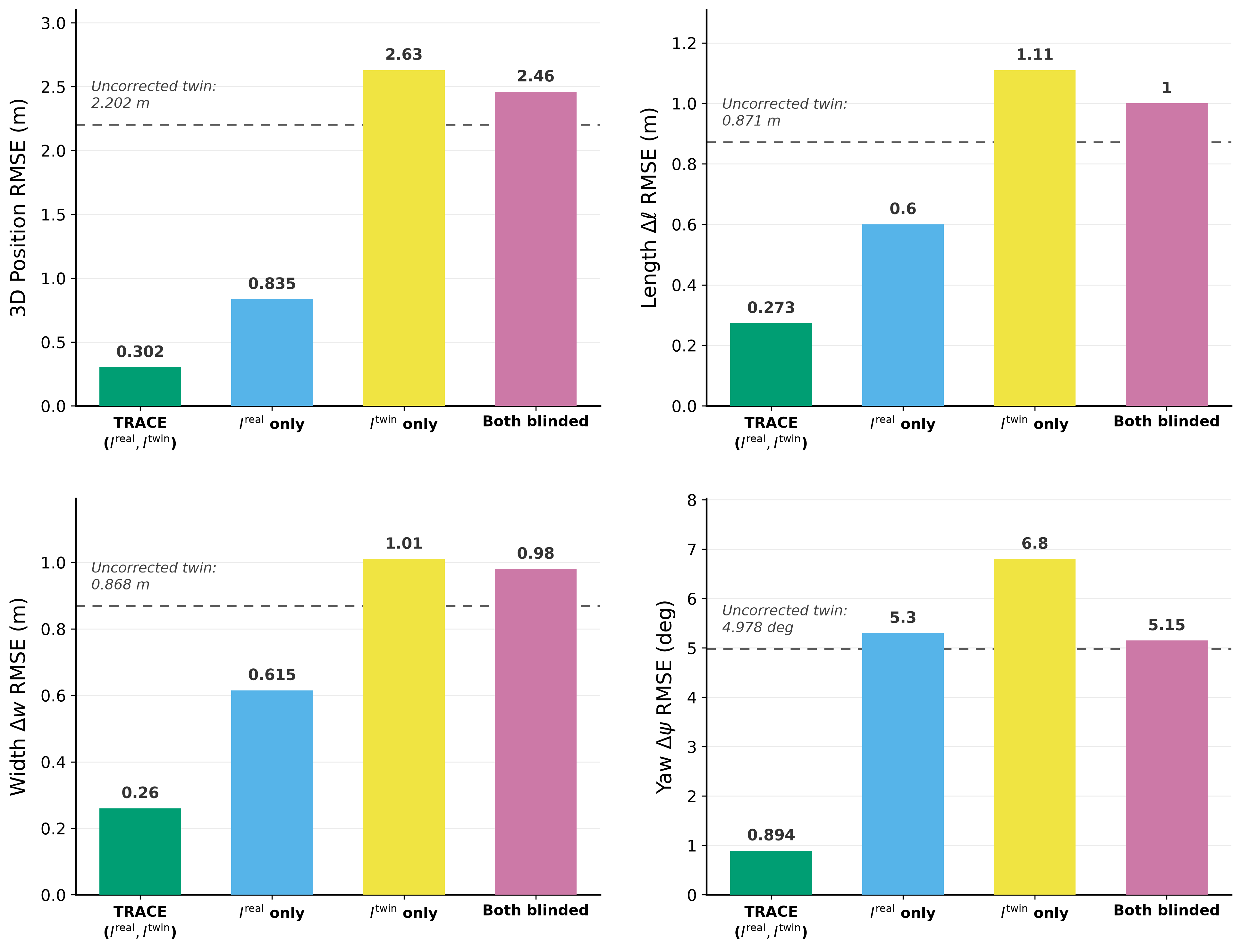}
    \caption{\textbf{Input blinding across geometric parameters.}
    The frozen model is evaluated with the full measured--rendered pair, measured RF only, twin-rendered RF only, and both RF branches blinded. Dashed lines show the corresponding error of the corrupted twin before
    correction. The 3D position result is reported in Sec.~\ref{sec:results_falsification}; the remaining panels show footprint length, footprint width, and yaw.}
    \label{fig:falsification_full}
\end{figure}

Figure~\ref{fig:falsification_full} shows that the same dependence on the paired RF evidence extends beyond position. With both RF branches, footprint length, footprint width, and yaw RMSE are \SI{0.273}{m},
\SI{0.260}{m}, and $0.894^\circ$, respectively. With measured RF alone, these increase to \SI{0.600}{m}, \SI{0.615}{m}, and $5.30^\circ$. With twin-rendered RF alone, the errors rise further to \SI{1.11}{m}, \SI{1.01}{m}, and $6.80^\circ$.

The degradation is especially strong for footprint dimensions and yaw. Measured RF alone still contains information about the physical structure, but without the matched twin rendering the model loses the RF reference describing what the current geometry predicts. Conversely, the twin-rendered branch alone contains the current hypothesis but no independent observation of the physical scene. This also explains why the twin-only and both-blinded cases can be worse than leaving the corrupted twin unchanged. The no-correction baseline applies no geometric update, whereas the frozen corrector still produces a residual under the blinded inputs. In the twin-only case, both $\vect{I}^{\twin}$ and $\vect{P}^{(t)}$ describe the same corrupted geometry, so there is no independent evidence indicating how that geometry should change. When both RF branches are blinded, only the current twin prior $\vect{P}^{(t)}$ remains. In either case, the predicted residual is no longer grounded by physical RF evidence and can therefore increase the geometric error. These additional results support the same conclusion as the 3D-position experiment: the correction is most accurate when the model can compare the physical RF evidence directly with the RF response predicted by the current twin.

\section{Out-of-Distribution Error and Closed-Loop Correction}
\label{app:closedloop}

This appendix evaluates \trace{} under two out-of-distribution settings, each of which moves one axis of the scene generator (App.~\ref{app:data}) outside its training support while holding the other fixed. In OOD-Layout, the planar components of $\vect{\eta}_i$ are drawn from $[-8,-6]\cup[6,8]$~m, outside the training layout support, while the twin corruption remains in distribution. In OOD-Error, the layout remains in distribution, but the corruption is drawn from $\varepsilon_i\in[-6,-4]\cup[4,6]$~m. Every corrupted building therefore requires a correction larger than any seen during training, where displacements are limited to $\pm$\SI{4}{m}. OOD-Layout asks whether the correction rule depends on \emph{where} buildings are; OOD-Error asks whether it depends on \emph{how wrong} the twin is. Table~\ref{tab:ood_extrapolation} summarizes both.

\begin{table}[h]
\centering
\caption{\textbf{Performance under out-of-distribution layout and error shifts.}
Fraction of initial 3D position RMSE removed when either the scene layout or the initial twin error lies outside the training support; higher is better. \trace{} step 2 applies the same frozen corrector after updating and re-rendering the twin from step 1.}
\label{tab:ood_extrapolation}
\small
\setlength{\tabcolsep}{7pt}
\begin{tabular}{lcc}
\toprule
\textbf{Method} & \textbf{OOD-Layout} & \textbf{OOD-Error} \\
 & \textbf{Removed [\%]} & \textbf{Removed [\%]} \\
\midrule
ViT               & 23.59 & 10.60 \\
U-Net             & 56.33 & 54.59 \\
\trace{}, step 1  & \textbf{88.70} & 79.16 \\
\trace{}, step 2  & --    & \textbf{93.08} \\
\bottomrule
\end{tabular}
\end{table}

\paragraph{\trace{} remains accurate beyond the training distribution.}
Under OOD-Layout, \trace{} removes 88.70\% of the initial 3D position error, versus 56.33\% for the U-Net and 23.59\% for the ViT. This is nearly unchanged from the in-distribution representation result in Table~\ref{tab:repr}. This robustness comes from GeoCrop: rather than processing the full scene, where the model could memorize the global layout, \trace{} compares the measured and rendered RF only within a local window around each building's current twin position to infer the geometric mismatch.



\paragraph{OOD-Error and closed-loop correction.}
OOD-Error is intentionally more challenging because the required geometric correction exceeds the corruption range seen during training. The regression head is therefore asked to extrapolate to residual magnitudes on which it was not optimized. \trace{} nevertheless removes 79.16\% of the initial error in a single pass, compared with 54.59\% for the U-Net and 10.60\% for the ViT. Because every displacement exceeds the training range, part of the largest corrections remains after this first pass. We address this residual by applying the same frozen corrector again to the updated twin, as defined by the update rule in Eq.~\eqref{eq:loop}.

\paragraph{The iteration.}
Each pass consists of the following steps:
\begin{enumerate}[leftmargin=1.6em,nosep]
  \item predict $g_i$ and $\vect{r}_i$ from $(\vect{I}^{\real},\vect{I}^{\twin},\vect{P}^{(t)})$;
  \item write the correction back to the twin, $\vect{p}_i^{(t+1)}=\vect{p}_i^{(t)}+g_i\vect{r}_i$;
  \item re-render the updated twin under the same acquisition configuration, $\vect{H}^{\twin}_{s}=\R(\vect{P}^{(t+1)};\vect{\xi}_s)$, and backproject it to obtain a new $\vect{I}^{\twin}$;
  \item recompute the GeoCrops, now centered on the updated twin locations, and apply the same frozen corrector again.
\end{enumerate}

\paragraph{Why the second pass succeeds.}
The first pass brings the twin closer to the physical scene. When the GeoCrops are re-centered on the updated building positions, the measured and rendered responses inside each window differ only by the remaining error, which is smaller and closer to the range seen during training. The same frozen corrector therefore operates on an easier problem, and error removed increases from 79.16\% to 93.08\%. No retraining or fine-tuning is involved; the improvement comes entirely from re-rendering the corrected twin.

\paragraph{What the loop requires.}
Only the twin branch is re-rendered. The measured branch is acquired once and reused across passes, so each additional pass costs one ray-tracing and backprojection of the current twin per sensing node, with no new measurement. Nothing in the loop requires the true geometry $\vect{P}^{\star}$, so the same procedure is directly applicable in deployment.


\section{Measured NIST Benchmark}
\label{app:nist}

Section~\ref{sec:results_real} reports the measured-RF result. This appendix documents the protocol behind it: the site, the two roles of LiDAR, how the training data was produced, and how the two branches are constructed at test time.

\begin{figure}[htbp]
    \centering
    \begin{subfigure}[b]{0.31\textwidth}
        \centering
        \includegraphics[width=\linewidth]{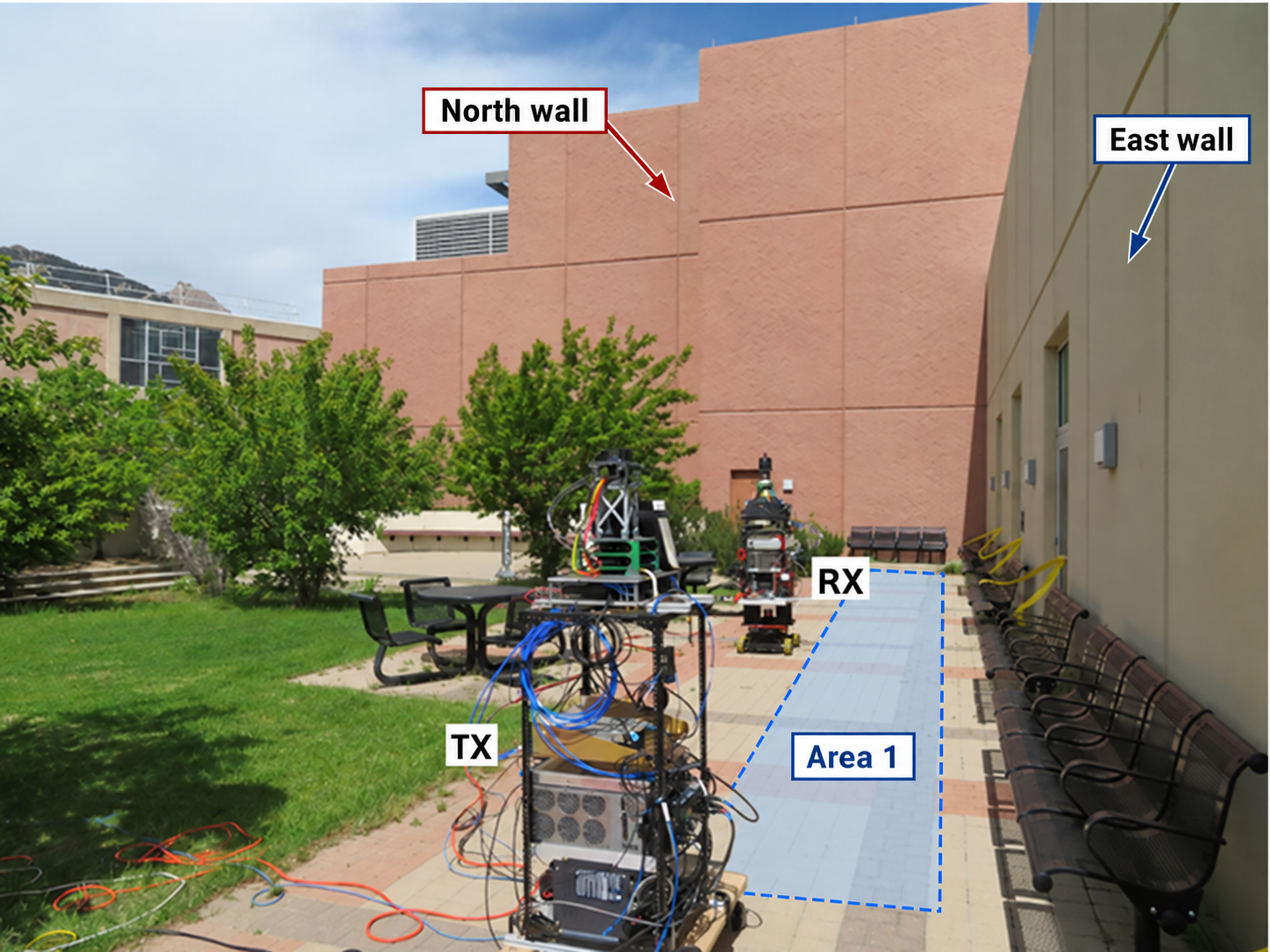}
        \caption{}
    \end{subfigure}\hfill
    \begin{subfigure}[b]{0.335\textwidth}
        \centering
        \includegraphics[width=\linewidth]{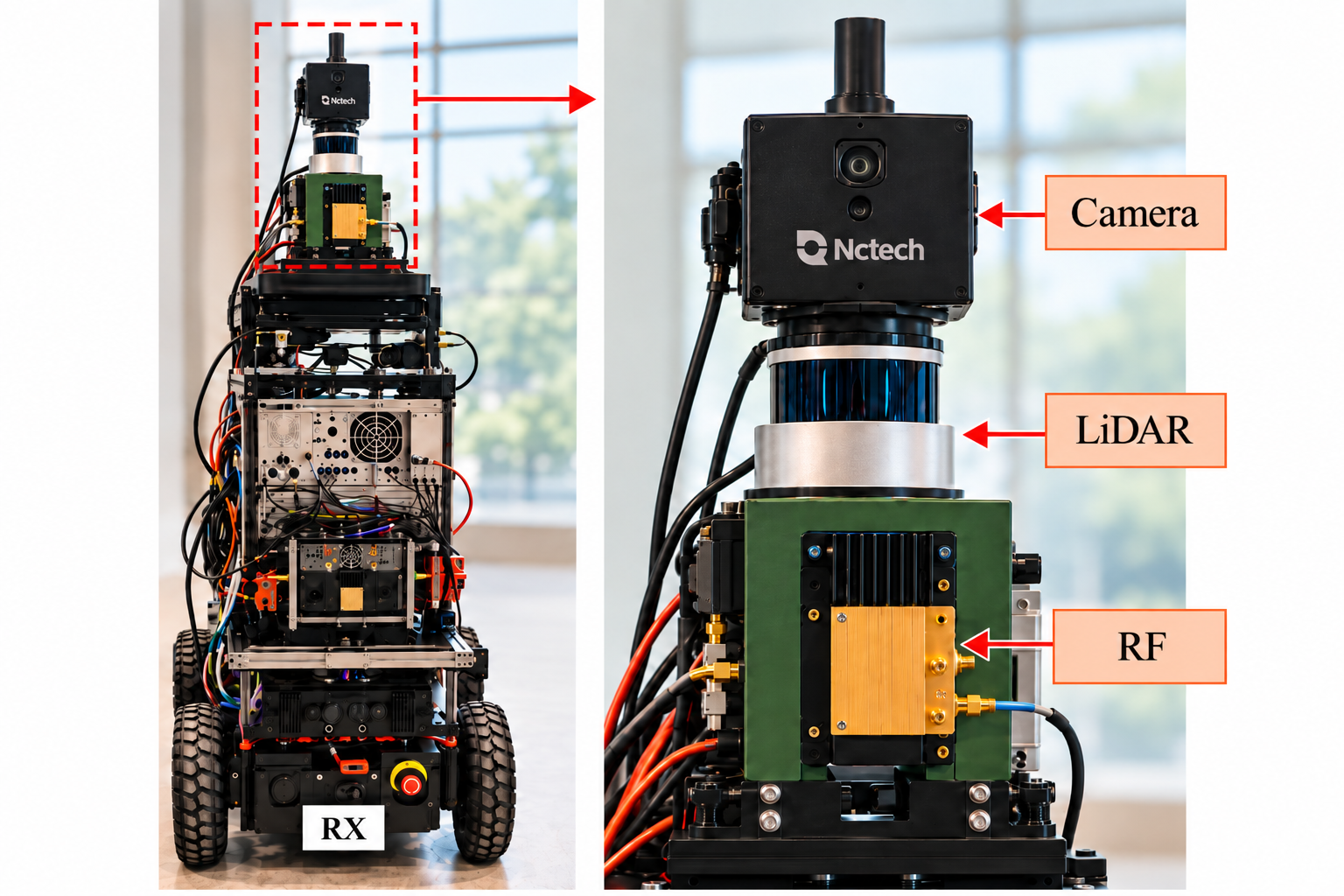}
        \caption{}
    \end{subfigure}\hfill
    \begin{subfigure}[b]{0.335\textwidth}
        \centering
        \includegraphics[width=\linewidth]{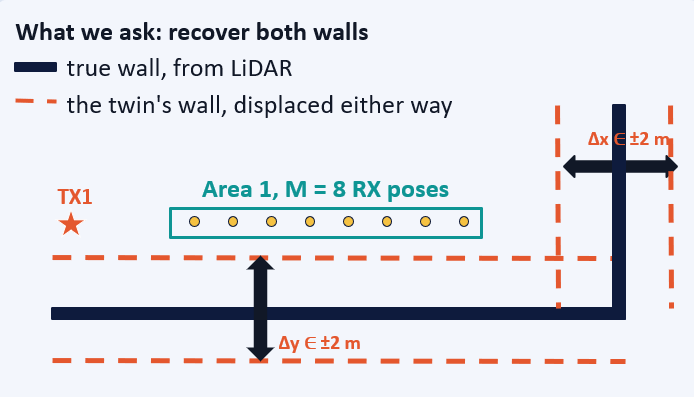}
        \caption{}
    \end{subfigure}
    \caption{\textbf{Measured-data benchmark, NIST outdoor courtyard \cite{nist_nextg_repository}.}
    (a) Measurement setup with fixed TX and a mobile RX region. (b) Mobile RX platform carrying RF, LiDAR, and camera sensors. (c) The planar recovery task: $M=8$ RX poses sampled in Area~1 with displaced twin walls. LiDAR provides the reference wall geometry; \trace{} is trained only on synthetic twin corruptions and evaluated on measured \SI{28}{GHz} RF.}
    \label{fig:nist_measured_data}
\end{figure}

\paragraph{Site and acquisition.}
The campaign is an outdoor courtyard measurement at \SI{28}{GHz} \cite{nist_nextg_repository}, collected with one fixed transmitter and a mobile receiver platform carrying RF, LiDAR, and camera sensors (Fig.~\ref{fig:nist_measured_data}). We use transmitter TX1 and a receiver strip in Area~1, and each evaluation group fuses $M=8$ receiver position measurements into a single estimate. Here $M$ is a physical quantity, the number of distinct receiver positions contributing to one estimate, and not a model hyperparameter; the corresponding closest quantity in simulation is the number of active sensing nodes. 


\paragraph{The two roles of LiDAR, and what the model receives.}
LiDAR is used twice, although in neither case does it reach the corrector. First, the LiDAR point cloud is the source of the reference site model from which the Sionna courtyard twin is built. Second, the LiDAR-derived wall geometry is the
ground truth against which the recovered wall positions are scored. The learned corrector itself receives RF evidence and the current twin state, as in simulation. The measured result is therefore RF-driven correction evaluated against a LiDAR reference, not RF-plus-LiDAR fusion.

\paragraph{Training data.}
From the LiDAR-derived site model we build a Sionna twin of the courtyard and use that twin to generate all training data. During training, the model sees only synthetic measured-to-rendered pairs, produced by rendering the reference courtyard twin as the physical branch and randomized corruptions of it as the twin branch.

\paragraph{Test-time construction.}
At test time the two branches are produced differently from each other. $\vect{I}^{\real}$ comes from the physical NIST RF measurement.
$\vect{I}^{\twin}$ is produced by ray tracing the corrupted courtyard twin in Sionna using the same transmitter and receiver configuration as that measurement, so the matched-acquisition condition of
App.~\ref{app:formulation} is preserved across the sim-to-real boundary. The corruption displaces two courtyard walls independently in the horizontal plane by up to $\pm\SI{2}{m}$, with the remaining geometry held at the reference, and a given sample may move one wall substantially while leaving the other near its reference position. The frozen corrector then predicts the planar wall corrections from the measured and rendered pair.

\begin{figure}[t]
    \centering
    \includegraphics[width=0.60\linewidth]{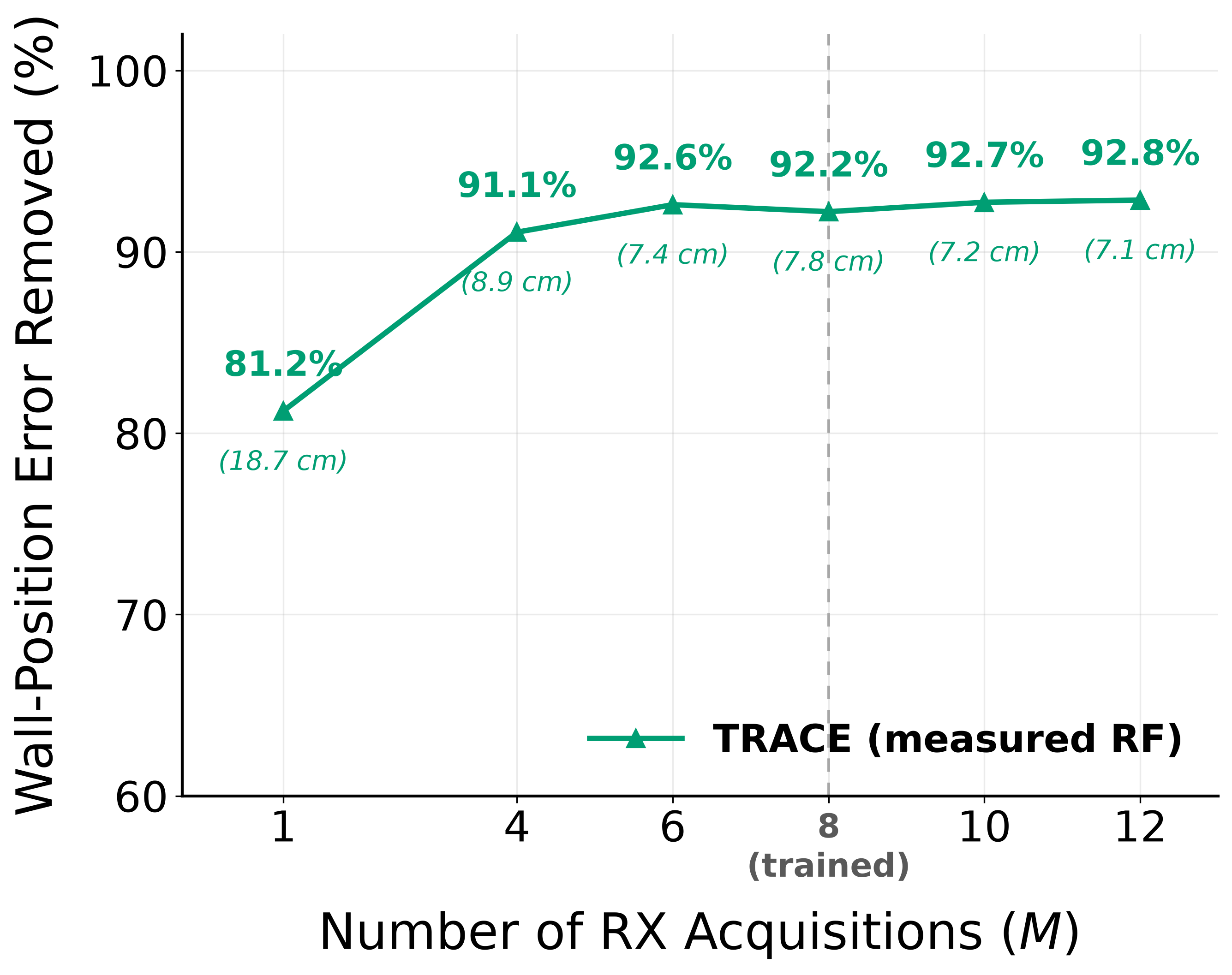}

    \caption{\textbf{Robustness to the number of measured receiver positions.} Fraction of the initial wall-position error removed as the number of receiver positions $M$ fused into each estimate varies; the mean planar error is given in parentheses. The same frozen checkpoint, trained with $M=8$, is evaluated without retraining. The uncorrected twin has \SI{1.00}{m} mean error.}
    \label{fig:nist_views}
\end{figure}

\paragraph{Robustness to the number of measured receiver poses.}
The main measured-RF result uses $M=8$ receiver positions per estimate, matching the configuration used during training. To test sensitivity to the number of available measured views, we evaluate the same frozen checkpoint without retraining for $M\in\{1,4,6,8,10,12\}$, where $M$ denotes the number of distinct receiver positions fused into one estimate. As shown in Fig.~\ref{fig:nist_views}, even a single receiver position reduces the mean planar wall-position error from \SI{1.00}{m} to \SI{18.7}{cm}, removing 81.2\% of the initial error. With $M=4$, the error decreases to \SI{8.9}{cm}, and for $M=6$--$12$ it remains near \SI{7}{cm}--\SI{8}{cm}. The results show that \trace{} remains effective with fewer measured views than used during training, while most of the additional gain is obtained from the first few independent receiver viewpoints.

\paragraph{Scope of the measured benchmark.}
In the measured benchmark, we evaluate planar wall-position correction; wall height and the remaining geometric attributes are held at the reference, while the full six-parameter correction is evaluated in simulation. No measured sample is used in training or fine-tuning.    
Unlike the synthetic benchmark, the measured-to-rendered residual is not
guaranteed to arise only from geometry: it may also contain material,
antenna, synchronization, clutter, and other modeling mismatch. The measured
experiment therefore tests whether the geometric corrector remains useful in
the presence of these unmodeled residual components. Because synthetic
training uses the NIST courtyard twin, this experiment evaluates
synthetic-to-measured RF transfer at a known site rather than cross-site
generalization.


\section{Implementation, Training, and Baseline Details}
\label{app:implementation}

This appendix records the configuration used to produce the reported numbers, grouped in pipeline order: scene and sensing generation, the deterministic front end, the learned corrector, the training and evaluation protocol, and
the baselines. Derived quantities are marked $^\dagger$.

\subsection{Synthetic scene and sensing configuration}
\label{app:data}

\begin{table}[htbp]
\caption{Sensing configuration and synthetic scene distribution.}
\label{tab:sensing}
\centering
\small
\setlength{\tabcolsep}{2.5pt}
\begin{tabular}{llll}
\toprule
Quantity & Value & Quantity & Value \\
\midrule
Carrier frequency          & \SI{28}{GHz}         & Sensing nodes $S$        & 3, monostatic \\
Bandwidth                  & \SI{122.88}{MHz}     & Node height              & \SI{15.5}{m} (rooftop) \\
TX array                   & $8\times8$           & Node downtilt            & \SI{12}{\degree} \\
RX array                   & $8\times8$           & Node aim point          & $(30, 0, 5)$~m \\
Ray tracer                 & Sionna RT            & Sensing-node building           & $8\times8\times\SI{15}{m}$ \\
\midrule
\multicolumn{4}{l}{\emph{Stage 1: true-scene building parameters}} \\
Height $h$                 & $[7, 15]$~m          & Length $\ell$            & $[8, 14]$~m \\
Width $w$                  & $[8, 14]$~m          & Yaw $\psi$               & $[0, 180)$\si{\degree} \\
Layout jitter $\eta_{\max}$ in $(x,y)$ & $\pm\SI{6}{m}$ & Overlapping layouts & rejected \\
\midrule
\multicolumn{4}{l}{\emph{Stage 2: twin corruption, training distribution}} \\
$\Delta x, \Delta y$       & $\pm\SI{4}{m}$       & $\Delta h$               & $\pm\SI{4}{m}$ \\
$\Delta \ell, \Delta w$    & $\pm\SI{3}{m}$       & $\Delta \psi$            & $\pm\SI{15}{\degree}$ \\
Corrupted per scene        & 1 to 3, uniform      & Remaining buildings      & left exact \\
\midrule
\multicolumn{4}{l}{\emph{Out-of-distribution evaluation sets}} \\
OOD-Layout, $\vect{\eta}_i$       & $[-8,-6]\cup[6,8]$~m & OOD-Layout, $\varepsilon_i$ & in-distribution, $[-4,4]$~m \\
OOD-Error, $\vect{\varepsilon}_i$ & $[-6,-4]\cup[4,6]$~m & OOD-Error, $\vect{\eta}_i$      & in-distribution, $[-6,6]$~m \\
\bottomrule
\end{tabular}
\end{table}


Table~\ref{tab:sensing} lists the sensing configuration and the two-stage scene generator. Scene generation has two independent stages, and keeping them independent is what allows layout diversity and correction difficulty to be swept separately. Stage 1 draws the physical scene by perturbing every building of a nominal template per parameter, $\vect{p}^{\star}_i=\vect{p}^{\rm nom}_i+\vect{\eta}_i$, so no two scenes share a layout and the nominal template is never itself a sample. Stage 2 forms the current twin by corrupting a random subset of one to three buildings, $\vect{p}^{(0)}_i=\vect{p}^{\star}_i+\vect{\varepsilon}_i$,
and leaving the rest exact. Both stages reject layouts with overlapping footprints or with footprints outside the imaging grid. Because the corruption defines the supervision target directly, $\Delta\vect{p}_i^{(0)}=-\vect{\varepsilon}_i$, and every scene retains uncorrupted buildings.

The two OOD sets each move one of these axes outside its training support while holding the other in distribution, which is what makes them separable diagnostics: OOD-Layout changes where the buildings are, and OOD-Error changes how wrong the twin is.

Receiver noise is injected only into the synthetic physical branch, and through a precomputed backprojected noise bank rather than by adding noise to the CSI, so that the noise field retains the spatial correlation that focusing
induces.

\subsection{World-frame backprojection and GeoCrop}

\begin{table}[htbp]
\caption{World-frame backprojection grid and GeoCrop.}
\label{tab:frontend}
\centering
\small
\setlength{\tabcolsep}{3.5pt}
\begin{tabular}{llll}
\toprule
Quantity & Value & Quantity & Value \\
\midrule
Grid $N_x \times N_y$      & $260 \times 198$ px  & Grid resolution          & \SI{0.4066}{m} / px \\
Extent in $x$              & $[-25.0, 80.3]$~m    & Extent in $y$            & $[-40.0, 40.1]$~m \\
Focus heights $K$          & 6                    & Focus height values      & $z = 0,3,6,9,12,15$~m \\
GeoCrop size               & $56 \times 56$ px    & GeoCrop extent$^\dagger$ & \SI{22.8}{m} square \\
Image scaling              & per scene, shared    & Scaling region           & town crop, all heights \\
\bottomrule
\end{tabular}
\end{table}

Table~\ref{tab:frontend} gives the world grid, the focus planes, and the crop. The GeoCrop window is \SI{22.8}{m} square. This is large enough to contain a building at the upper end of the generated footprint range together with a planar displacement at the upper end of the training corruption range, so the
structure the model must align remains inside the window in the axis-aligned case, and largely so for rotated buildings. The per-scene intensity scale is computed once over a fixed region of the
world grid covering the built area, listed as the scaling region in Table~\ref{tab:frontend}, and applied identically to the measured, rendered, and coherent-difference streams. This
is deliberate: normalizing the two branches separately would remove the relative brightness difference between them, which is part of the evidence, and normalizing per focus height would remove the height-dependent contrast that the vertical stack exists to expose.

\subsection{Learned corrector}

\begin{table}[htbp]
\caption{Architecture of the learned \trace{} corrector $f_\theta$.}
\label{tab:architecture}
\centering
\small
\setlength{\tabcolsep}{2.5pt}
\begin{tabular}{llll}
\toprule
Quantity & Value & Quantity & Value \\
\midrule
Token dimension $D_{\mathrm{emb}}$ & 256                  & Tokens per crop $T^\dagger$   & $7 \times 7 = 49$ \\
Complex CNN stages         & 4                            & CNN channels (complex)        & $6/16/32/64/128$ \\
Cross-attention            & 1 layer, 4 heads             & CueFusion projection          & $5D_{\mathrm{emb}} \to D_{\mathrm{emb}}$ \\
Sensor fusion              & 2 layers, 4 heads            & Building attention            & 3 layers, 4 heads \\
FiLM initialization        & zero (identity at start)     & Dropout                       & 0.2 \\
Parameters (full \trace{}) & 3{,}979{,}911 (3.98\,M)      & Absolute $(x,y)$ in graph     & excluded \\
\bottomrule
\end{tabular}
\end{table}

One encoder is shared across all buildings, all sensing nodes, and all three
branches, which is why the parameter count is independent of $N$ and $S$. The
$6$ input channels of the first complex stage are the $K=6$ focus heights. The
CueFusion projection is $5D_{\rm emb}\to D_{\rm emb}$ because the coherent
residual stream is enabled; the ablation row that removes it reduces this to
four cues.

\subsection{Training and evaluation protocol}
\label{app:training}

\begin{table}[htbp]
\caption{Training and evaluation settings.}
\label{tab:training}
\centering
\small
\setlength{\tabcolsep}{3.5pt}
\begin{tabular}{llll}
\toprule
Quantity & Value & Quantity & Value \\
\midrule
\multicolumn{4}{l}{\emph{Optimization}} \\
Optimizer                  & AdamW                    & Learning rate           & $1 \times 10^{-4}$ \\
Weight decay               & $1 \times 10^{-3}$       & Batch size              & 128 \\
Schedule                   & ReduceLROnPlateau        & Warmup                  & 8 epochs \\
Max epochs                 & 200                      & Early-stop patience     & 20 epochs \\
Gradient clipping          & 1.0 (max norm)           & Mixed precision         & enabled \\
\midrule
\multicolumn{4}{l}{\emph{Loss and targets}} \\
$\ell_{\mathrm{reg}}$      & MSE                      & $\lambda_{\mathrm{reg}} / \lambda_{\mathrm{gate}}$ & $1.0 / 0.25$ \\
Offset normalization       & per dimension            & Yaw representation      & double angle \\
Gate at evaluation         & soft (sigmoid)           &                         & \\
\midrule
\multicolumn{4}{l}{\emph{Noise and data}} \\
Training SNR               & \SI{15}{dB}              & Evaluation SNR          & $-5$ to \SI{20}{dB} \\
Noise branch               & physical only            & Evaluation noise        & resampled per run \\
Sensor dropout             & $p = 0.5$                & Building dropout        & disabled \\
Train / val / test         & $0.70 / 0.15 / 0.15$     & Split level             & scene disjoint \\
Held-out samples           & 5{,}400                  & Random seed             & 42 \\
\bottomrule
\end{tabular}
\end{table}

Splits are disjoint at the level of the underlying true scene, so no test
layout appears during training. Checkpoints are selected on a
dimension-balanced validation score rather than on a single aggregate
distance, because the six residual dimensions have different physical scales
and an aggregate distance is dominated by position; a checkpoint chosen on
that aggregate would be selected almost entirely on $(\Delta x,\Delta y)$.

Evaluation noise is resampled per run, which is what the protocol of
App.~\ref{app:ablation} accounts for when differencing the ablation costs.

\subsection{Baselines}
\label{app:baselines}

Both full-scene baselines consume the complete backprojected scene, with
sensing nodes, focus heights, and the three streams stacked as ordinary
real-valued channels, giving the $S\times3\times2\times K=108$ input channels
of Table~\ref{tab:baselines}; the factor of two separates real and imaginary
parts, since neither baseline is complex-valued. The input size in
Table~\ref{tab:baselines} is written as image rows by columns,
$198\times260$, which is the same grid as the $N_x\times N_y=260\times198$ of
Table~\ref{tab:frontend} with the world axes in the opposite order. They inherit the dataset, the
scene-disjoint split, the normalization, the noise model, the residual
targets, the move-gated loss of Eq.~\eqref{eq:loss}, the optimizer, the
schedule, the validation score, and the evaluation metrics described above.
The comparison therefore isolates the input representation and the fusion
strategy rather than the training recipe.


\begin{table}[htbp]
\caption{Configurations of the ViT and U-Net baselines.}
\label{tab:baselines}
\centering
\small
\setlength{\tabcolsep}{3.5pt}
\begin{tabular}{llll}
\toprule
Quantity & Value & Quantity & Value \\
\midrule
\multicolumn{4}{l}{\emph{Shared}} \\
Input channels$^\dagger$   & $S \times 3 \times 2 \times K = 108$ & Input size      & $198 \times 260$ \\
Head trunk                 & $256 + 6 \to 256$          & Heads                   & $g_i$, $\vect{r}_i$ (6) \\
Missing sensors            & zero-filled channels     & Sub-pixel correction    & not applicable \\
\midrule
\multicolumn{4}{l}{\emph{U-Net}} \\
Base channels              & 44                       & Depth                   & 3 enc / 3 dec, bottleneck \\
Feature dimension          & 256                      & Pooling                 & multi-scale adaptive \\
Trainable parameters       & 3{,}922{,}123 (3.92\,M)  &                         &  \\
\midrule
\multicolumn{4}{l}{\emph{ViT}} \\
Patch size                 & 7 px                     & Tokens$^\dagger$        & $29 \times 38 = 1102$ \\
Embedding dimension        & 232                      & Depth / heads           & 4 / 8 \\
MLP ratio                  & 4.0                      & Positional embedding    & learned, absolute \\
Trainable parameters       & 4{,}143{,}679 (4.14\,M)  &                         &  \\
\bottomrule
\end{tabular}
\end{table}

The two baselines differ in how they build a global scene descriptor. The U-Net
encodes the backprojected scene with a convolutional encoder-decoder and pools
multi-scale features, so it retains local spatial structure; it is the
stronger of the two on this task (Table~\ref{tab:main}); each building's prediction is then read out from the
shared descriptor together with its current state through the head trunk. The
ViT partitions the same input into $7\times7$-pixel patches, adds a learned
absolute positional embedding, and attends globally, so its token
representation is tied to position within the global image. It is also the
weaker of the two under the layout shift of Table~\ref{tab:ood_extrapolation},
removing 23.59\% of the initial error against 56.33\% for the U-Net.

The U-Net and ViT contain 3{,}922{,}123 (3.92\,M) and 4{,}143{,}679 (4.14\,M) trainable parameters, respectively, compared with 3{,}979{,}911 (3.98\,M) for \trace{}. All three models are therefore matched in capacity to within about 4\%, so the performance differences are not attributable to a larger learned model. The ablations support the same conclusion: the smallest \trace{} variant, with 2.93\,M parameters after removing cross-sensor attention (Table~\ref{tab:ablation}), still removes 83.0\% of the initial 3D error, against 61.6\% for the U-Net.




\end{document}